\documentclass[journal,twocolumn]{IEEEtran}

\usepackage{amsmath}
\usepackage{amssymb}
\usepackage{epsfig, psfrag}
\usepackage{amsthm}
\usepackage{latexsym}
\usepackage{bbm}
\usepackage[ruled,lined]{algorithm2e}
\SetKw{Init}{Initialize:}
\SetKw{Def}{Definition:}

\allowdisplaybreaks

\usepackage{epstopdf}
\usepackage{subfigure}
\usepackage{graphics,color}
\usepackage{graphicx}

\usepackage{mathtools}
\mathtoolsset{showonlyrefs=true}

\usepackage{hyperref}

\newtheorem{theorem}{Theorem}
\newtheorem{lemma}{Lemma}

\newtheorem{fact}{Fact}
\theoremstyle{definition}

\newtheorem{remarks}{Remark}

\def\expe{\mathbb{E}}   
\def\P{\mathsf{P}}   
\def\q{\mathsf{Q}}
\def\argmin{\mathop{\rm arg\,min}}
\def\argmax{\mathop{\rm arg\,max}}

\def\mc{\mathcal}

\def\mbb{\mathbb}

\newcommand{\indicate}[1]{\mathbf{1}{\{#1\}}}

\begin{document}


\title{Bayesian Algorithms for Decentralized Stochastic Bandits}
\author{\IEEEauthorblockN{Anusha Lalitha$^1$, and Andrea Goldsmith$^{1,2}$}\\
\IEEEauthorblockA{$^1$ Stanford University, $^2$ Princeton University \\
Email: alalitha@stanford.edu, goldsmith@princeton.edu}
\thanks{Part of this paper is under review in~\cite{dects_aistats}.}  \thanks{The work of A.~Lalitha and A.~Goldsmith was supported by ONR Grant N00014-18-1-2191.}
}


\maketitle

\begin{abstract}
    We study a decentralized cooperative multi-agent multi-armed bandit problem with $K$ arms and $N$ agents connected over a network. In our model, each arm's reward distribution is same for all agents, and rewards are drawn independently across agents and over time steps. In each round, agents choose an arm to play and subsequently send a message to their neighbors. The goal is to minimize cumulative regret averaged over the entire network. We propose a decentralized Bayesian multi-armed bandit framework that extends single-agent Bayesian bandit algorithms to the decentralized setting. Specifically, we study an information assimilation algorithm that can be combined with existing Bayesian algorithms, and using this, we propose a decentralized Thompson Sampling algorithm and decentralized Bayes-UCB algorithm. We analyze the decentralized Thompson Sampling algorithm under Bernoulli rewards and establish a problem-dependent upper bound on the cumulative regret. We show that regret incurred scales logarithmically over the time horizon with constants that match those of an optimal centralized agent with access to all observations across the network. Our analysis also characterizes the cumulative regret in terms of the network structure. Through extensive numerical studies, we show that our extensions of Thompson Sampling and Bayes-UCB incur lesser cumulative regret than the state-of-art algorithms inspired by the Upper Confidence Bound algorithm.  We implement our proposed decentralized Thompson Sampling under gossip protocol, and over time-varying networks, where each communication link has a fixed probability of failure.
\end{abstract}

\section{Introduction}
Stochastic Multi-armed bandits (MABs) is a fundamental and well-studied model for sequential decision making.  Here, an agent repeatedly chooses an arm from a finite collection of arms, each of which produces a reward drawn from a fixed but unknown distribution associated with each arm. MAB models the exploration/exploitation trade-off inherent in sequential decision problems. At every instant, the agent must choose whether to use the observations already gathered to gain the greatest immediate reward (exploitation) or select an unexplored arm to gain greater knowledge and potential future reward (exploration). Decision-making problems also arise in decentralized contexts, such as a fully decentralized recommendation system, where content is recommended based on user feedback without running through a central server~(\cite{pmlr-v28-szorenyi13, 10.1145/2911451.2911548, 6709807, 10.1016/j.chb.2014.12.011, sankararaman2019social}). Another example is real-time traffic planning using a decentralized sensor network in which agents try to optimize a route in a shared environment~\cite{pmlr-v28-szorenyi13}.

Motivated by the growing need to design learning algorithms for large scale networked systems, we consider a collaborative multi-agent decentralized extension of the classical stochastic MAB problem. We consider a network of $N$ agents playing the same instance of a $K$-armed MAB problem.  In this work, agents aim to minimize the average cumulative regret over the network, i.e., per-agent cumulative regret in the network. In our model, agents can send messages over a communication network connecting them. Thus, agents can potentially collaborate to speed up learning by interacting with each other rather than learning in isolation.  

The literature on stochastic MAB algorithms at a single agent can be separated into two distinct categories: frequentist algorithms and Bayesian algorithms. Using a frequentist algorithm, such as the Upper Confidence Bound (UCB) algorithm, the agent maintains an empirical estimate of the mean rewards and a confidence interval around it. The agent uses the sum of the empirical mean of the rewards observed and confidence interval to choose an arm at each time instant.  In contrast, in the Bayesian algorithms such as Thompson Sampling~\cite{10.1093/biomet/25.3-4.285} and Bayes-UCB~\cite{pmlr-v22-kaufmann12}, each arm's reward distribution is characterized by a parameter which is endowed with a prior distribution. The agent maintains a posterior distribution over reward distributions' parameters and exploits the knowledge of the whole posterior to select an arm at each time. Bayesian algorithms such as Thompson Sampling have better empirical performance than the UCB algorithm and achieve optimal regret bounds~\cite{pmlr-v31-agrawal13a}. However, prior work has focused only on extending frequentist algorithms, specifically the UCB algorithm, to the decentralized MAB problem. Our simulations show that when agents implement Thompson Sampling for certain problem instances and network topologies, they incur lesser regret than some of the previously proposed decentralized frequentist algorithms even if they do not interact with others (Figures~\ref{fig:dducb_langdren_decTS} and \ref{fig:dducb_langdren_decTS_bayesucb}). Thus, our objective is to propose a generic decentralized bandit framework to extend single-agent Bayesian MAB algorithms and their benefits to the decentralized MAB problem.

\subsection{Contributions}
In this paper, we design and analyze decentralized Bayesian MAB algorithms. Our contributions are as follows:

\begin{itemize}
    \item 
    We propose a decentralized Bayesian MAB framework that extends single-agent Bayesian MAB algorithms to a decentralized setup. We use this to design a decentralized Thompson Sampling algorithm and a decentralized BayesUCB algorithm. 
    
    \item
    We show that for Bernoulli rewards, our proposed decentralized Thompson Sampling achieves asymptotically optimal per-agent regret that matches the performance of a centralized agent with full cooperation. Furthermore, for decentralized Thompson Sampling, we establish a problem-dependent upper bound on the per-agent cumulative regret that scales logarithmically in time horizon and characterizes network structure dependence. 
    
    \item
    We provide extensive simulations for Gaussian and Bernoulli bandits to show that our algorithms outperform state-of-the art algorithms for decentralized MAB problem. We also provide extensive numerical studies showing the effect of network topology. 
    
    \item
    We implement our proposed decentralized Thompson Sampling under the gossip protocol and over time-varying networks, where each communication link has a fixed probability of failure. We demonstrate that, surprisingly, decentralized Thompson Sampling achieves logarithmic regret under time-varying networks.
\end{itemize}

\subsection{Related Work}

In recent years, the multi-agent stochastic MAB setting has received increasing attention.  Several works have studied multi-agent stochastic MAB problems but with different formulations and communication models. 

The formulation we consider in this work, where $N$ agents are playing the same instance of a $K$-armed bandit, has been studied previously by Mart{\'{\i}}nez{-}Rubio et~al.~\cite{nips2019} and Landgren et~al.\cite{landgren_cdc16}. Decentralized frequentist algorithms proposed by these works are inspired from the well-known UCB algorithm. In these algorithms, each agent relies on a running consensus algorithm to estimate the mean rewards from its local rewards and its neighbors' estimated rewards. Each agent then applies the UCB algorithm using the values estimated through the running consensus algorithm. Landgren et~al.~\cite{landgren_cdc16} propose three different algorithms for decentralized MAB inspired from UCB: coop-UCB, coop-UCB2, and coop-UCL, where coop-UCL is a decentralized Bayesian algorithm obtained by extending single agent Bayesian algorithm the Upper Credible Limit (UCL) algorithm propoosed by Reverdy et~al.~\cite{6774467}. While coop-UCB2 and coop-UCL only require the knowledge of the number of agents, coop-UCB also requires the knowledge of the graph's spectral gap and the whole set of eigenvectors of the communication matrix. Furthermore, coop-UCB requires tuning of the parameter $\gamma$ in a problem-specific manner to achieve optimal performance. When implementing the coop-UCL algorithm, every agent maintains a posterior distribution over the mean reward for all arms. Agents obtain posterior distributions by plugging in approximate frequentist estimators for the mean. This work generalizes the coop-UCL algorithm by proposing a decentralized Bayesian MAB framework to extend single-agent Bayesian algorithms by directly combining the posterior distributions for any parametric family of reward distributions without relying on frequentist estimates.

Mart{\'{\i}}nez{-}Rubi et~al.~\cite{nips2019} propose the Decentralized Delayed Upper Confidence Bound algorithm (DDUCB), which utilizes a variant of the running consensus algorithm known as Chebyshev acceleration to accelerate the information assimilation over the network. However, DDUCB also requires knowledge of the number of agents in the network and an upper bound on the spectral gap of the communication matrix. In contrast to both these works, the proposed algorithm decentralized Thompson Sampling only requires knowledge of the number of agents in the network, and does not have any parameters to tune. We show in our simulations that decentralized Thompson Sampling achieves a significant reduction in the regret incurred over these algorithms. Moreover, our simulations show that for certain problem instances and network topologies, the coop-UCB algorithm and the DDUCB algorithm incur a greater regret than when agents implement Thompson Sampling in isolation. These results demonstrate that Bayesian algorithms such as Thompson Sampling can not only lead to superior performance in centralized settings, but also in a decentralized setup.  

The earliest works in distributed multi-agent MAB setting are by Awerbuch et~al.~\cite{10.1007/11503415_16} and Cesa-Bianchi et~al.~\cite{10.5555/3322706.3322723}, however for the adversarial bandits. This was followed by studies on multi-agent stochastic MAB peer-to-peer networks where the underlying graph structure is fully connected. Korda et~al.~\cite{pmlr-v48-korda16} consider a problem where all the agents are trying to solve the same underlying linear bandit problem and send information to any other agent in the network picked via a gossip protocol. Similarly, Szorenyi~\cite{pmlr-v28-szorenyi13} also considers the MAB problem in P2P random networks. The trade-off between communication cost and regret minimization among team of bandits has been studied by several works~\cite{7218651, ijcai2017-24, sankararaman2019social}. Chakraborty et~al.~\cite{ijcai2017-24} consider the setting where agents can broadcast their last obtained reward every agent and forgo pulling an arm or playing an arm and collecting a reward. This is an extreme setup where communication is expensive and even a centralized algorithm achieves no-regret. Furthermore, in these models, when agents communicate they simultaneously share information with all others.

Sankararaman et~al.~\cite{sankararaman2019social} consider a setting where agents make arm recommendations over a social network. Their work aims to propagate the best arm effectively via a gossip protocol while requiring $O(\log T)$ communication rounds. However, arm recommendations can reduce the regret only when agents are not aware of arms locally and rely on recommendations from neighbors to learn about the optimal arm. Hillel et~al.~\cite{DBLP:conf/nips/HillelKKLS13} study the communication versus simple regret, i.e., pure exploration for best arm identification, which is different from the cumulative regret, i.e., the explore-exploit trade-off considered in this paper. There is another line of work in distributed settings where the agents are competitive (\cite{5535151, 6763073, 10.5555/3327757.3327824}). Liu and Zhao~\cite{5535151} consider a distributed MAB problem where agents do not exchange information however if there is a collision i.e., two agents pull the same arm, the reward is arbitrarily split or no reward is obtained at all. Kalathil et~al.~\cite{6763073} and Nayyar~\cite{7765076} consider a similar collision model with communication that increases the regret incurred.   Shahrampour et~al.~\cite{shahrampour_dece_MAB} consider a decentralized MAB problem where agents aim to maximize the sum of rewards of all agents across the network. However, they assume the existence of a centralized controller that selects actions for the agents.

Recently Madhushani and Leonard in~\cite{madhushani2019heterogeneous} analyze a decentralized UCB algorithm over a network with stochastic communication links. They show that agents incur logarithmic regret when agents can observe the actions and rewards of their neighbors. In this work, we empirically show that our proposed decentralized Thompson Sampling algorithm also incurs logarithmic regret. In our algorithms, agents merge their local posteriors using a merge rule that has been used for information assimilation in distributed hypothesis testing and estimation~\cite{7349151, 7172262, dht_tit_2018}. To the best of our knowledge, our work is the first to analyze the posterior merge rule when applied to sequential decision-making agents under partial feedback such as stochastic MABs, and to show that logarithmic regret is achievable in this setting. We refer the reader to~\cite{dht_tit_2018} and related references for other applications of the posterior merging rule. This works builds upon our previous work~\cite{dects_aistats} where we proposed the decentralized Thompson Sampling algorithm and its regret analysis. This paper provides a more generalized approach to the decentralized Bayesian MABs ~\cite{dects_aistats}, and proposes an additional algorithm, the decentralized Bayes-UCB. We also provide additional numerical simulations where a decentralized MAB algorithm, is shown to incur logarithmic regret over time-varying networks.  

\underline{Notation:} For any positive integer $T$, let $[T]:= \{1,\ldots, T\}$. The Kullback--Leibler (KL) divergence between two probability density functions $p_1(\cdot)$ and $p_2(\cdot)$ with respect to measure $\mu$ on space $\mathcal{X}$ is defined as $D_{KL}(p_1\parallel p_2)=\int_{\mathcal{X}} p_1(x) \log\frac{p_1(x)}{p_2(x)} d\mu(x)$, with the convention $0 \log \frac{a}{0}=0$ and $b \log \frac{b}{0}=\infty$ for $a,b\in [0,1]$ with $b\neq 0$.

\section{Problem Formulation}
\label{sec:prob_formulation}
Consider an MAB problem with $N$ agents connected through an undirected connected graph $\mc{G}$. Each agent $i \in [N]$ sequentially chooses arms $\{A^{(i)}_{t}\}_{t \in \mbb{N}}$ from a finite set of arms $\mc{A}$, where $|\mc{A}| = K < \infty$. The set of arms $\mc{A}$ is same for all agents in the network. At every agent $i$, there is a random reward $Y^{(i)}_{t, k} \in \mc{Y}$ associated with each arm $k \in [K]$ and time $t \in \mbb{N}$. We consider a parametric family of distributions $\mc{P} = \{p_{\theta}\}_{\theta \in \Theta}$ where each $p_{\theta}$ is a probability distribution over $\mc{Y}$. When agent $i$ plays an arm $k$ at time $t$, it observes a reward $Y^{(i)}_{t, k}$ sampled from the distribution $p_{\theta^{\ast}_k}$ independent of arms played at previous time steps and of arms played by other agents. Also, rewards associated with any arm come from the same distribution common to all agents in the network. At each time instant, agents are allowed to send messages to their neighbors as dictated by the underlying graph $\mc{G}$ and, similarly, receive messages from their respective neighbors to use it in the next time step.



We are interested in the frequentist setting where $\{\theta^{\ast}_k\}_{k \in [K]}$ (the problem instance) is fixed. Let $\mu_k := \expe_{\theta^{\ast}_k}[Y^{(i)}_{t,k}]$ and assume without loss of generality that $\mu_1 \geq \mu_2 \geq \ldots \mu_K$. Let the sub-optimality gap be defined as $\Delta_k := \mu_1 - \mu_k$ and define $\Delta := \min_{a \neq 1}\Delta_a = \Delta_2$. We also assume that the true underlying parameters $\{\theta^{\ast}_k\}_{k \in [K]}$ or expected values of rewards $\{\mu_k\}_{k \in [K]}$ are unknown to the agents. All agents
play the same $K$-armed bandit problem for $T$ time steps and aim to minimize the per-agent regret over the network $R(T)$ defined as
\begin{align}
    R(T)
    :=
    \frac{1}{N}\sum_{i=1}^{N}\expe_{\theta^{\ast}}\left[\sum_{t=1}^T \left(Y^{(i)}_{t, A_1} - Y^{(i)}_{t, A^{(i)}_t}\right)\right],
\end{align}
where $\theta^{\ast} = [\theta^{\ast}_1, \ldots, \theta^{\ast}_K]$  and the expectation is taken over the randomness in the arms played $A^{(i)}_t$ and rewards $Y^{(i)}_{t}$. The collective goal for each agent is to minimize the per-agent regret while exchanging at most $\mathsf{poly}(K)$ messages with their neighbors per iteration.

If the agents do not interact at all, then each agent will see a standard $K$-armed MAB problem. Applying the well known Upper Confidence Bound (UCB) algorithm, we obtain that per agent regret after playing for $T$ time steps scales as $O(\frac{K}{\Delta}\log T)$~\cite{lattimore_szepesvari_2020}. On the other extreme is the optimal centralized model where we assume perfect collaboration. Each agent plays an arm, observes a reward, and then broadcasts both the arm played and the reward obtained to all other agents at every time step. After $T$ time steps, each agent has reward equivalent to having played $NT$ number of arms. Again applying the UCB algorithm, we can see that the per agent regret scales as $O(\frac{K}{N\Delta}\log NT) = O(\frac{K}{N\Delta}\log T + \frac{\log N}{N\Delta})$. This is the best possible per-agent regret scaling one can obtain in a networked setting using UCB~\cite{lattimore_szepesvari_2020}. We aim to design a decentralized arm selection strategy for a network of $N$ agents playing the same MAB problem, which achieves a per agent regret close to the one incurred by the optimal centralized algorithm while exchanging at most $\mathsf{poly}(K)$ messages with their neighbors per iteration.


\section{Decentralized Bayesian MAB Algorithms}
In this section, we present a framework to extend any single agent MAB Bayesian algorithm to the decentralized MAB formulation discussed in Section~\ref{sec:prob_formulation}. Then, we apply our framework to extend two well-known Bayesian bandit algorithms namely Thompson Sampling and Bayes-UCB.

We assume that agents take a Bayesian approach. Specifically, each agent assumes a prior distribution on the parameters $\theta^{\ast}_k$ that parametrize the reward distribution for arm $k \in [K] $. Bayesian MAB algorithms maintain a posterior distribution for each parameter $\theta^{\ast}_k$ for every arm $k \in [K] $ using the past rewards and the prior distribution. At every time step, each agent plays an arm according to its local Bayesian MAB algorithm which is a function of the posterior probability of the parameters $\{\theta^{\ast}_k\}_{k \in [K]}$. After observing the local reward, each agent updates the local posterior distribution associated with parameters of the selected arm. Then, the agents exchange the posterior probability distribution of all parameters $\{\theta^{\ast}_k\}_{k \in [K]}$. Agents combine the posterior probabilities by taking an average of the log of the posterior distributions. They use the merged posterior distributions to select an arm in their next play.

Let $q_0$ be the Radon–Nikodym derivative of a prior $Q_0$ with respect to some measure $\mu$ on $\Theta$. For every $k \in [K]$, let $Q^{(i)}_{k,t}(\cdot)$ be the posterior distribution of $\theta^{\ast}_k$ maintained by agent $i$ after $t$ rounds of playing and communication with its neighbors. Let $q^{(i)}_{k,t}(\cdot)$ denote its density with respect to measure $\mu$. Let $W \in \mbb{R}^{N \times N}$ denote a communication matrix of non-negative weights that respects the structure of the graph $\mc{G}$, meaning that for all $i \neq j$, $W_{ij} > 0$ if and only if there exists an edge between agent $i$ and agent $j$. Furthermore, we assume that $W$ is doubly stochastic.  In the social learning literature~\cite{7349151, 7172262, dht_tit_2018}, the value $W_{ij}$ captures the weight that agent $i$ assigns to messages from agent $j$. However, in this work (as we will show later), introducing a communication matrix $W$ provides the agents a simple and effective way to merge the posterior distributions of each $\theta^{\ast}_k$ maintained by each agent. Let $\mc{N}(i):= \{j \in [N]: W_{ij}> 0\}$ denote the neighborhood of agent $i$. From here onward, to simplify our notation we use $Y^{(i)}_t$ to denote the reward observed by agent $i$ at time $t$ for arm $A^{(i)}_t$. Now, we are ready to provide our decentralized Bayesian learning framework for multi-agent MAB networks, as captured in Algorithm~\ref{alg:decent_mab}. 

\begin{algorithm}[h]
    \DontPrintSemicolon
    \KwIn{initial prior $q_0$, learning rate $\eta >0$, communication matrix $W$}
    \Init $q^{(i)}_{k,1} \gets q_0$ for all $i \in [N]$ and $k \in [K]$ \;
    \For{$t = 1,2, \ldots$}{
        \For{$i = 1, \ldots, N$}{
            Select arm $A^{(i)}_t \gets$ \texttt{Bayes\_MAB}$\left(q^{(i)}_{1,t}, \ldots, q^{(i)}_{K,t}\right)$\;
            Play $A^{(i)}_t$ and Observe $Y^{(i)}_{t} \sim p_{\theta^{\ast}_{A^{(i)}_t}}$\;
            Update posterior distribution:\;
            $\tilde{q}^{(i)}_{A^{(i)}_t,t+1}(\theta) \gets \frac{q^{(i)}_{A^{(i)}_t,t}(\theta)p^{\eta}_{\theta}\left(Y^{(i)}_{t}\right)}{\int_{\phi \in \Theta}q^{(i)}_{A^{(i)}_t,t}(\phi)p^{\eta}_{\phi}\left(Y^{(i)}_{t}\right)d \mu(\phi)}, \, \forall \theta \in \Theta$\;
            Send messages $\{\tilde{q}^{(i)}_{k, t+1}\}_{k \in [K]}$ to all $j \in \mc{N}(i)$\;
        }
        \For{$i = 1, \ldots, N$}{
            \For{$k = 1, \ldots, K$}{
                Merge posteriors for all $\theta \in \Theta$:\;
                $q^{(i)}_{k,t+1}(\theta) \gets \frac{\exp\left(\sum_{j=1}^{N} W_{ij}\log \tilde{q}^{(j)}_{k,t+1}(\theta)\right)}{\int_{\phi \in \Theta}\exp\left(\sum_{j=1}^{N} W_{ij}\log \tilde{q}^{(j)}_{k,t+1}(\phi)\right)d\mu(\phi)}$
            }
        }
    }
\caption{Decentralized Bayesian MAB Algorithm}
\label{alg:decent_mab}
\end{algorithm}

Algorithm~\ref{alg:decent_mab} provides a framework to extend any single agent Bayesian MAB algorithm, denoted by \texttt{Bayes\_MAB}, to our decentralized MAB setup. At each step, agent $i$ selects an arm as dictated by the \texttt{Bayes\_MAB} algorithm and observes a reward for the selected arm. Then, the agent $i$ updates its posterior distribution using a variation of Bayes rule to obtained a \textit{tempered posterior distribution} as given below
\begin{align}
    \label{eq:posterior_update}
    \tilde{q}^{(i)}_{A^{(i)}_t,t+1}(\theta) = \frac{q^{(i)}_{A^{(i)}_t,t}(\theta)p^{\eta}_{\theta}\left(Y^{(i)}_{t}\right)}{\int_{\phi \in \Theta}q^{(i)}_{A^{(i)}_t,t}(\phi)p^{\eta}_{\phi}\left(Y^{(i)}_{t}\right)d \mu(\phi)},
\end{align}
where $\eta$ is positive constant. Such tempered posteriors have been well studied in the literature~\cite{Grnwald2012TheSB, alquier2020}. Tempered posteriors are robust to miss-specified models~\cite{Grnwald2012TheSB}. For a single-agent, $\eta$ is chosen to be less than $1$. For our decentralized setting, our analysis shows that $\eta = N$ guarantees asymptotically optimal regret bounds.

Every agent $i$ exchanges the tempered posterior distributions for all arms of its neighbors $j \in \mc{N}(i)$. Following this, each agent $i$ takes a weighted average of the log tempered posterior distributions from its neighbors $j \in \mc{N}(i)$, where the weights are given by the $i$-th row of the communication matrix $W$. Specifically, the merged posterior for arm $k$ at each agent is given by
\begin{align}
    \label{eq:merge_posteriors}
    q^{(i)}_{k,t+1}(\theta) = \frac{\exp\left(\sum_{j=1}^{N} W_{ij}\log \tilde{q}^{(j)}_{k,t+1}(\theta)\right)}{\int_{\phi \in \Theta}\exp\left(\sum_{j=1}^{N} W_{ij}\log \tilde{q}^{(j)}_{k,t+1}(\phi)\right)d\mu(\phi)},
\end{align}
for all $\theta \in \Theta$. When $A^{(i)}_t = k$ for some arm $k$, then the above update algorithm for arm $k$ can be written as
\begin{align}
q^{(i)}_{k,t+1}(\theta)
= 
\frac{p^{\eta}_{\theta}(Y^{(i)}_{t} )\prod_{j=1}^N(q^{(j)}_{k,t}(\theta))^{W_{ij}}}{\int_{\phi \in {\Theta}}p^{\eta}_{\phi}(Y^{(i)}_{t} )\prod_{j=1}^N(q^{(j)}_{k,t}(\phi))^{W_{ij}}d\mu(\phi)}.
\end{align}
Furthermore, in this case the posterior distribution $q^{(i)}_{k,t+1}(\theta)$ can be obtained as the closed form solution of the following decentralized optimization problem
\begin{align}
&\argmin_{\pi} \left\{ \expe_{\theta \sim \pi}[-\log p_{\theta}(Y^{(i)}_{t}  )] 
+ \frac{1}{\eta}\sum_{j=1}^N W_{ij} D_{KL}(\pi \parallel q^{(j)}_{k,t})\right\},
\end{align}
where $\eta$ can be interpreted as a learning rate~\cite{7349151, 7172262}.

Implementing Algorithm~\ref{alg:decent_mab} may require additional tools from the Bayesian computational toolbox to perform an exact or approximate posterior update given by equation~\eqref{eq:posterior_update} and to compute the exact or appropriate merged posterior given by equation~\eqref{eq:merge_posteriors}. We now look at two commonly studied models where equations~\eqref{eq:posterior_update} and~\eqref{eq:merge_posteriors} can be implemented
exactly without resorting to numerical approximation.

\begin{lemma}[Beta-Bernoulli Bandits]
\label{lemma:beta_bernoulli_update}
Suppose the rewards are sampled from the family of Bernoulli distributions and the prior on the Bernoulli parameters at each agent $i\in [N]$ for arm $k\in [K]$ is assumed to be $\mathsf{Beta}(\alpha_0, \beta_0)$. The posterior update at agent $i$ at time $t \geq 1$ given by equation~\eqref{eq:posterior_update} is given as
\begin{align}
    \tilde{\alpha}^{(i)}_k(t+1) &=  \alpha^{(i)}_k(t) +\eta Y^{(i)}_{t} \indicate{A^{(i)}_t = k},
    \\
    \tilde{\beta}^{(i)}_k(t+1) &=  \beta^{(i)}_k(t) +\eta(1- Y^{(i)}_{t}) \indicate{A^{(i)}_t = k},
\end{align}
and we have $\tilde{Q}^{(i)}_{k,t+1} = \mathsf{Beta}\left(\tilde{\alpha}^{(i)}_k(t+1), \tilde{\beta}^{(i)}_k(t+1) \right), \forall i \in [N], k \in [K]$. Furthermore, the posterior merge at agent $i$ at time $t \geq 1$ given by equation~\eqref{eq:merge_posteriors} is given as
\begin{align}
\label{eq:alpha_merge}
\alpha^{(i)}_k(t+1) &= \sum_{j=1}^N W_{ij} 
\tilde{\alpha}^{(j)}_k(t+1),\\ 
\label{eq:beta_merge}
\beta^{(i)}_k(t+1) &= \sum_{j=1}^N W_{ij} 
\tilde{\beta}^{(j)}_k(t+1),
\end{align}
and we have $Q^{(i)}_{k,t+1} = \mathsf{Beta}(\alpha^{(i)}_{k,t+1}, \beta^{(i)}_{k,t+1}), \forall i \in [N], k \in [K]$.
\end{lemma}

\begin{lemma}[Gaussian Bandits]
\label{lemma:gaussian_update}
Suppose the rewards are sampled from the family of Gaussian distributions with variance $\sigma^2$ and the prior on the mean value at each agent $i\in [N]$ for arm $k\in [K]$ is assumed to be $\mc{N}(\mu_0, \sigma^2_0)$. The posterior update at agent $i$ at time $t \geq 1$ given by equation~\eqref{eq:posterior_update} is given as
\begin{align}
    \tilde{\sigma}^{(i)}_k(t+1) &=  \sqrt{\frac{1}{\frac{1}{(\sigma^{(i)}_{k}(t))^2} + \frac{\eta \indicate{A^{(i)}_t = k}}{\sigma^2}}},
    \\
    \tilde{\mu}^{(i)}_k(t+1) &=  \mu^{(i)}_{k}(t)\left(\frac{\tilde{\sigma}^{(i)}_k(t+1)}{\sigma^{(i)}_k(t)}\right)^2 +
    \nonumber
    \\
    & + \eta Y^{(i)}_t\indicate{A^{(i)}_t = k}\left(\frac{\tilde{\sigma}^{(i)}_k(t+1)}{\sigma}\right)^2, 
\end{align}
and we have $\tilde{Q}^{(i)}_{k,t+1} = \mc{N}\left(\tilde{\mu}^{(i)}_k(t+1), (\tilde{\sigma}^{(i)}_k(t+1))^2 \right), \forall i \in [N], k \in [K]$. Furthermore, the posterior merge at agent $i$ at time $t \geq 1$ given by equation~\eqref{eq:merge_posteriors} is given as
\begin{align}
\label{eq:sigma_merge}
\frac{1}{(\sigma^{(i)}_k(t+1))^{2}} &= \sum_{j=1}^N W_{ij} 
\frac{1}{(\tilde{\sigma}^{(j)}_k(t+1))^{2}},\\ 
\label{eq:mu_merge}
\frac{\mu^{(i)}_k(t+1)}{(\sigma^{(i)}_k(t+1))^{2}} &= \sum_{j=1}^N W_{ij} \frac{\tilde{\mu}^{(j)}_k(t+1)}{(\tilde{\sigma}^{(i)}_k(t+1))^{2}},
\end{align}
and we have $Q^{(i)}_{k,t+1} = \mc{N}\left(\mu^{(i)}_k(t+1), (\sigma^{(i)}_k(t+1))^2 \right), \forall i \in [N], k \in [K]$.
\end{lemma}

Lemma~\ref{lemma:beta_bernoulli_update} and Lemma~\ref{lemma:gaussian_update} are proved in  Appendix~\ref{sec:additional_lemmata}.

In the Bayesian MAB literature, a prior is chosen over the reward distribution parameters such that it is a conjugate prior to the family of reward distributions. This ensures that the posterior distributions are in the same probability distribution family as the prior probability distribution. Hence, for Bernoulli bandits we choose a prior from the family of Beta distributions. For Gaussian bandits with known variance we choose a prior from the family of Gaussian distributions.  
Lemma~\ref{lemma:beta_bernoulli_update} and Lemma~\ref{lemma:gaussian_update} show that under Algorithm~\ref{alg:decent_mab}, when we choose a prior that is a conjugate prior to the reward distribution we are still guaranteed that the posterior distributions are in the same probability distribution family as the prior when rewards come from Bernoulli distribution or Gaussian distributions. Similar result can be obtained for reward distributions that belong to a one-parameter exponential family. 

The algebraic convenience of choosing a conjugate prior in the posterior update step extends to our posterior merge step as well, since we can obtain a closed-form expression for the merged posterior; otherwise, numerical integration may be necessary. In the case of Bernoulli bandits, merging the posteriors is equivalent to taking a simple convex combination of the shape parameters of the Beta posteriors, namely, alpha and beta parameters given by equation~\eqref{eq:alpha_merge} and~\eqref{eq:beta_merge}. Similarly for Gaussian bandits, merging the posteriors is equivalent to taking a simple convex combination of the natural parameters of the Gaussian posteriors as given by equation~\eqref{eq:sigma_merge} and~\eqref{eq:mu_merge}. Therefore, introducing a doubly stochastic communication matrix $W$ provides a simple and effective way to keep a running approximation of the average of parameters of the posterior distributions across all agents in the network. This is closely connected to the running consensus algorithm used by frequentist algorithms for our decentralized MAB problem for assimilation of information across the network. For instance, Mart{\'{\i}}nez{-}Rubio et~al.~\cite{nips2019} and Landgren et~al.\cite{landgren_cdc16}, propose extensions of UCB to decentralized setup which leverage running consensus algorithm and its variants to obtain accurate approximations of the number of times arm $k$ was pulled and sum of rewards coming from all the pulls done to arm $k$ across the network.

\subsection{Decentralized Thompson Sampling}

The decentralized Bayesian MAB algorithm can be used to extend Thompson Sampling to the decentralized MAB problem discussed in Section~\ref{sec:prob_formulation}. The Thompson Sampling algorithm is a Bayesian algorithm which relies on drawing samples from the posterior distributions $\{q^{(i)}_{k,t}\}_{k \in [K]}$ in order to select each arm $k$ with a probability equal to the probability that its mean $\mu_k$ is the highest~\cite{10.1093/biomet/25.3-4.285}. In other words, an agent implementing Thompson Sampling selects an arm according to its posterior probability of being the best arm. After observing the reward, the agent updates its posterior distributions $\{q^{(i)}_{k,t}\}_{k \in [K]}$. Combining this with Algorithm~\ref{alg:decent_mab} gives us the decentralized Thompson Sampling algorithm described in Algorithm~\ref{alg:decent_ts}. In decentralized Thompson Sampling algorithm, each agent selects an arm by implementing the Thompson Sampling algorithm as a function of its local posteriors. After observing the reward, each agent updates its posterior according to equation~\eqref{eq:posterior_update}. Then, the agents send their local updated posteriors to their neighbors. Each agent combines the posteriors received from its neighbors according to equation~\eqref{eq:merge_posteriors}. Agents use the merged posterior to select an arm in their next play.  

\begin{algorithm}[h]
    \DontPrintSemicolon
    \KwIn{initial prior $q_0$, learning rate $\eta >0$, communication matrix $W$}
    \Init $q^{(i)}_{k,1} \gets q_0$ for all $i \in [N]$ and $k \in [K]$ \;
    \For{$t = 1,2, \ldots$}{
        \For{$i = 1, \ldots, N$}{
            $\theta^{(i)}_{k,t} \sim q^{(i)}_{k, t}, \, \forall k \in [K]$\;
            Select arm $A^{(i)}_t \gets \argmax_{k \in [K]} \expe_{\theta^{(i)}_{k,t}}[Y^{(i)}_{t}]$\;
            Play $A^{(i)}_t$ and Observe $Y^{(i)}_{t} \sim p_{\theta^{\ast}_{A^{(i)}_t}}$\;
            Update posterior distribution:\;
            $\tilde{q}^{(i)}_{A^{(i)}_t,t+1}(\theta) \gets \frac{q^{(i)}_{A^{(i)}_t,t}(\theta)p^{\eta}_{\theta}\left(Y^{(i)}_{t}\right)}{\int_{\phi \in \Theta}q^{(i)}_{A^{(i)}_t,t}(\phi)p^{\eta}_{\phi}\left(Y^{(i)}_{t}\right)d \mu(\phi)}, \, \forall \theta \in \Theta$\;
            Send messages $\{\tilde{q}^{(i)}_{k, t+1}\}_{k \in [K]}$ to all $j \in \mc{N}(i)$\;
        }
        \For{$i = 1, \ldots, N$}{
            \For{$k = 1, \ldots, K$}{
                Merge posteriors for all $\theta \in \Theta$:\;
                $q^{(i)}_{k,t+1}(\theta) \gets \frac{\exp\left(\sum_{j=1}^{N} W_{ij}\log \tilde{q}^{(j)}_{k,t+1}(\theta)\right)}{\int_{\phi \in \Theta}\exp\left(\sum_{j=1}^{N} W_{ij}\log \tilde{q}^{(j)}_{k,t+1}(\phi)\right)d\mu(\phi)}$
            }
        }
    }
\caption{Decentralized Thompson Sampling}
\label{alg:decent_ts}
\end{algorithm}

\subsection{Decentralized Bayes-UCB}

Decentralized Bayesian MAB algorithm can be used to extend Bayes-UCB to the decentralized MAB problem discussed in Section~\ref{sec:prob_formulation}. The Bayes-UCB algorithm proposed by Kaufmann et~al.~\cite{pmlr-v22-kaufmann12} is a Bayesian algorithm which relies on quantiles of the posterior distributions. Specifically, Bayes-UCB algorithm utilizes posterior over the means $\{\mu_1, \ldots, \mu_K\}$, and let $\{\rho^{(i)}_{k,t}\}_{k \in [K]}$ denote the current posterior over the means induced by $\{q^{(i)}_{k,t}\}_{k \in [K]}$. At every time instant, the agent computes confidence indices from fixed-level quantiles of $\{q^{(i)}_{k,t}\}_{k \in [K]}$ where the quantile level is chosen to be of the order $1/t$. Each agent plays an arm which maximizes the confidence indices. After observing the reward, the agent updates its posterior distributions $\{q^{(i)}_{k,t}\}_{k \in [K]}$. Combining this with Algorithm~\ref{alg:decent_mab} gives us the decentralized Bayes-UCB algorithm described in Algorithm~\ref{alg:decent_bayesUCB}. In decentralized Bayes-UCB algorithm, agents select an action according to their local Bayes-UCB algorithm and update their posterior according to equation~\eqref{eq:posterior_update}. Then, the agents send their local updated posteriors to their neighbors. Each agent combines the posteriors received from its neighbors according to equation~\eqref{eq:merge_posteriors}. Agents use the merged posterior to select an arm in their next play.

\begin{algorithm}[h]
    \DontPrintSemicolon
    \KwIn{time horizon $T$, parameters of the quantile $c$, initial prior $q_0$, learning rate $\eta >0$, communication matrix $W$}
    \Def Denote $\{\rho^{(i)}_{k,t}\}_{k \in [K]}$ as posterior over means $[\mu_1, \ldots, \mu_K]$\;
    Denote \texttt{Quantile}$(t, \rho)$ as quantile function associated with distribution $\rho$ such that $\P_{\rho}(X \leq \texttt{Quantile}(t, \rho)) = t$\;
    \Init $q^{(i)}_{k,1} \gets q_0$ for all $i \in [N]$ and $k \in [K]$ \;
    \For{$t=1 , \ldots, T$}{
        \For{$i = 1, \ldots, N$}{
            \For{$k = 1, \ldots, K$}{
            Compute:\; 
            $C^{(i)}_k(t) \gets \texttt{Quantile}\left(1-\frac{1}{t(\log T)^c}, \rho^{(i)}_{k,t} \right)$ 
            }
            Select arm $A^{(i)}_t \gets \argmax_{k \in [K]} C^{(i)}_k(t)$\;
            Play $A^{(i)}_t$ and Observe $Y^{(i)}_{t} \sim p_{\theta^{\ast}_{A^{(i)}_t}}$\;
            Update posterior distribution:\;
            $\tilde{q}^{(i)}_{A^{(i)}_t,t+1}(\theta) \gets \frac{q^{(i)}_{A^{(i)}_t,t}(\theta)p^{\eta}_{\theta}\left(Y^{(i)}_{t}\right)}{\int_{\phi \in \Theta}q^{(i)}_{A^{(i)}_t,t}(\phi)p^{\eta}_{\phi}\left(Y^{(i)}_{t}\right)d \mu(\phi)}, \, \forall \theta \in \Theta$\;
            Send messages $\{\tilde{q}^{(i)}_{k, t+1}\}_{k \in [K]}$ to all $j \in \mc{N}(i)$\;
        }
        \For{$i = 1, \ldots, N$}{
            \For{$k = 1, \ldots, K$}{
                Merge posteriors for all $\theta \in \Theta$:\;
                $q^{(i)}_{k,t+1}(\theta) \gets \frac{\exp\left(\sum_{j=1}^{N} W_{ij}\log \tilde{q}^{(j)}_{k,t+1}(\theta)\right)}{\int_{\phi \in \Theta}\exp\left(\sum_{j=1}^{N} W_{ij}\log \tilde{q}^{(j)}_{k,t+1}(\phi)\right)d\mu(\phi)}$
            }
        }
    }
\caption{Decentralized Bayes-UCB}
\label{alg:decent_bayesUCB}
\end{algorithm}

\section{Regret Analysis}
In this section, we consider the case where rewards have Bernoulli distribution, and provide analytical guarantees for per-agent cumulative regret.

\begin{theorem}
\label{thm:regret_decTS}
Consider the decentralized multi-armed bandit problem with $N$ agents, $K$ arms and Bernoulli rewards. Let $W$ be a doubly stochastic communication matrix. For any $\epsilon > 0$, choosing $\eta = N$, and prior as Beta$(1, 1)$, i.e., uniform distribution, the per-agent cumulative regret achieved by decentralized Thompson Sampling after $T$ rounds of play can be upper bounded as 
\begin{align}
R(T)
&\leq
\sum_{k=2}^K \Delta_k(1+\epsilon)^2\frac{\log NT}{N d(\mu_k , \mu_1)}
\nonumber
\\
& +
\frac{3\left(1+\frac{8}{\epsilon}\right)\log N}{1-\lambda_2(W)}\sum_{k=2}^K \Delta_k
+ O\left(\frac{1}{\epsilon^{\tilde{N}}} \right),
\end{align}
where $d(a,b) = a \log \frac{a}{b} + (1-a)\log \frac{1-a}{1-b}$ denotes the KL-divergence between two Bernoulli distributions, $\lambda_2(W)$ denotes the second largest eigenvalue of matrix $W$ in absolute value  and $\tilde{N} = \frac{N\log N}{1-\lambda_2(W)}$.
\end{theorem}

\begin{remarks}[Optimality of Decentralized Thompson Sampling]
Asymptotically, the per-agent regret incurred by decentralized Thompson Sampling scales logarithmically with the time horizon $T$ which satisfies
\begin{align}
\lim_{T \to \infty} \frac{R(T)}{\log T} 
\leq \sum_{k=2}^K\frac{\Delta_k}{Nd(\mu_k, \mu_1)}.
\end{align}
We compare this with the regret incurred by an optimal centralized agent with perfect collaboration, where each agent plays an arm, observes a reward and broadcasts both the arm played and the reward observed to all other agents in the network. This implies that after $T$ times steps, each agent is equivalent to a centralized agent that plays an $NT$ number of arms. Using the lower bound shown by Lai and Robbins~\cite{lai_robbins}, the per-agent regret can be lower bounded as
\begin{align}
    \lim_{T \to \infty} \frac{R(T)}{\log T} 
    \geq \sum_{k=2}^K\frac{\Delta_k}{Nd(\mu_k, \mu_1)}.
\end{align}
Therefore, our bounds in Theorem~\ref{thm:regret_decTS} asymptotically achieve the optimal lower bounds of a centralized agent. For Bernoulli reward distribution, our decentralized Thompson Sampling is the first algorithm with an analytical guarantee that asymptotically matches the performance of a centralized agent with full cooperation.  
\end{remarks}

Theorem~\ref{thm:regret_decTS} has been proved in detail in Appendix~\ref{sec:main_proof}. Here we provide a proof sketch.

\textit{Proof sketch:} Analysis of the decentralized Thompson Sampling algorithm is inspired from the analysis of single agent Thompson Sampling by Agrawal and Goyal~\cite{pmlr-v31-agrawal13a}. To analyze the per-agent regret recall that 
\begin{align}
R(T)
& = \frac{1}{N} \sum_{k=2}^K\Delta_k \sum_{i=1}^N\expe[n_k^{(i)}(T)],
\end{align}
where $n^{(i)}_k(T)$ denotes the number of times a sub-optimal arm $k$ was played at agent $i$ up to time $T$\footnote{We drop writing $\theta^{\ast}$ explicitly whenever it is understood from the context.}. Furthermore, we can write
\begin{align}
    \sum_{i=1}^N\expe[n^{(i)}_k(T)]
    &= \sum_{i=1}^N\expe\left[\sum_{t=1}^T \indicate{A^{(i)}_t = k}\right].
\end{align}
At every time instant, we condition the event that a sub-optimal arm $k$ will be played at any agent in the network on certain good events which are defined similar to those in the analysis of Thompson Sampling in~\cite{pmlr-v31-agrawal13a}. Hence, for each arm $k \in [K]$, fix two thresholds $x_k$ and $y_k$ such that $\mu_k < x_k < y_k < \mu_1$. Define events
\begin{align}
	E^{(i)}_{k}(t) := \left\{\hat{\mu}^{(i)}_{k}(t)\leq x_k \right\},
\quad
	\widetilde{E}^{(i)}_{k}(t) 
	:= \left\{\theta^{(i)}_{k}(t)\leq y_k \right\},
\end{align}
where $\hat{\mu}^{(i)}_{k}(t)$ denotes the empirical mean value of arm $k$ at agent $i$ at time $t$. The events $E^{(i)}_{k}(t)$ and $\widetilde{E}^{(i)}_{k}(t)$ are good events where $\hat{\mu}^{(i)}_{k}(t)$ and $\hat{\mu}^{(i)}_{k}(t)$ are not too far from the true mean $\mu_k$. Following the intuition of single-agent Thompson Sampling, we aim to show that these events will hold with high probability for most time steps at all agents in the network. Hence, conditioning on these good events we obtain
\begin{align}
    \sum_{i=1}^N\expe[n^{(i)}_k(T)]
    &= \sum_{i=1}^N\expe\left[\sum_{t=1}^T\indicate{A^{(i)}_t = k, \overline{E^{(i)}_{k}(t)}}\right]
    \nonumber
    \\
    &+\sum_{i=1}^N\expe\left[\sum_{t=1}^T\indicate{A^{(i)}_t = k, E^{(i)}_{k}(t), \overline{\widetilde{E}^{(i)}_{k}(t)}}\right]
    \nonumber
    \\
    & + \sum_{i=1}^N\expe\left[\sum_{t=1}^T \indicate{A^{(i)}_t = k, E^{(i)}_{k}(t), \widetilde{E}^{(i)}_{k}(t)}\right]. 
    \label{eq:cond_good_events}
\end{align}
Set $\eta = N$ and for $\epsilon>0$, using the three main lemmata proved in Appendix~\ref{sec:main_lemmata}, we show that the first term on the right hand side in equation~\eqref{eq:cond_good_events} can be upper bounded as $O\left(\tilde{N} \right)$, the second term can be upper bounded as $O\left( (1+\frac{8}{\epsilon})\tilde{N} + (1+\epsilon)^2\frac{\log NT}{d(\mu_k,\mu_1)}\right)$ and the third term can be loosely upper bounded as $O\left(\frac{1}{\epsilon^{\tilde{N}}}\right)$.

\begin{remarks}[Comparison with previous work]
Landgren et~al.~\cite{landgren_cdc16} propose three different algorithms for decentralized MAB inspired from UCB: coop-UCB, coop-UCB2, and coop-UCL. Their algorithm coop-UCB requires the knowledge of the number of agents and the spectral gap, furthermore, it also requires the knowledge of the whole set of eigenvectors of the communication matrix $W$.  
However, the proposed algorithm ``decentralized Thompson Sampling" only requires the knowledge of the number of agents in the network. 
As shown in the simulations, it achieves a significant reduction in the regret incurred over these algorithms. 
Furthermore, coop-UCB requires tuning of the parameter $\gamma$ in a problem specific manner to achieve optimal performance, when $\eta = N$ our algorithm does not have any parameters to tune. Mart{\'{\i}}nez{-}Rubi et~al.~\cite{nips2019} propose Decentralized Delayed Upper Confidence Bound algorithm (DDUCB) which requires the knowledge of the number of agents in the network and an upper bound on the spectral gap of the communication matrix. 
Our algorithm, while requiring less information regarding the communication matrix $W$, significantly improves the per agent regret as seen in the simulations (Figures~\ref{fig:dducb_langdren_decTS} and \ref{fig:dducb_langdren_decTS_bayesucb}). 
Furthermore, unlike our algorithm, DDUCB also requires tuning of parameters in a problem specific manner to achieve optimal performance. 
\end{remarks}

\begin{remarks}[Benefits]
Thompson Sampling is a Bayesian algorithm which exploits the knowledge of the whole posterior to determine the next action. Our posterior merging rule given by equation~\eqref{eq:merge_posteriors} assimilates the information across the network such that after sufficient time steps of the order $O\left( \frac{N\log N}{1-\lambda_2(W)}\right)$, the posterior available at each agent is close to the posterior of a centralized agent that has access to all the observations in the network\footnote{This can be seen in Lemma~\ref{lemma:alpha_beta_bounds} for bounds on shape parameters of Beta posterior at each agent.}. This allows our decentralized Thompson Sampling algorithm to also utilize a posterior that is close to the posterior of a centralized agent. Exploiting the whole posterior that is close to the posterior of a centralized agent rather than estimates of rewards and number of times an arm was played in the network, allows our decentralized Thompson Sampling to reduce the regret it incurs over prior work significantly. While our analysis is limited to Bernoulli rewards, decentralized Thompson Sampling achieves superior performance for rewards with Gaussian distribution. 
\end{remarks}

\begin{remarks}[Limitations]
The upper bound on cumulative regret provided by Theorem~\ref{thm:regret_decTS} sheds light on how close the per-agent cumulative regret incurred by decentralized Thompson Sampling is to that of a centralized agent. Specifically, we see that as the number of agents $N$ increases or as the \textit{spectral gap} $1-\lambda_2(W)$ decreases, it takes longer for the posteriors across the network to merge and hence longer to converge to the asymptotic constant in front of $\log T$ for per-agent regret. Analysis of the frequentist algorithms for decentralized MAB proposed by Mart{\'{\i}}nez{-}Rubio et~al.~\cite{nips2019} and Landgren et~al.~\cite{landgren_cdc16} provides an upper bound on the per-agent regret where the terms independent of time horizon $T$ scale logarithmically in $N$ and inversely with the spectral gap $1-\lambda_2(W)$. Our current analysis of the per-agent regret incurred by decentralized Thompson Sampling does not provide such a tight bound in terms of $N$ and spectral gap $1-\lambda_2(W)$. However, our simulations show superior performance over all the previously proposed algorithms indicating that terms that are independent of $T$ are much smaller than what our current analysis predicts. Improving our analysis to obtain a tighter bound in terms of $N$ and $1-\lambda_2(W)$ is a topic for future work.  
\end{remarks}

\section{Simulations}

In this section, we present simulations comparing our proposed algorithms' performance with prior work and study the effect of the number of agents and network topology on per-agent cumulative regret. Furthermore, we simulate our decentralized Thompson Sampling algorithm under gossip protocol and over time-varying networks, where each communication link has a fixed probability of failing. The communication matrix $W$ for all our simulations is doubly stochastic and designed according to Section~3.2 in~\cite{duchi_dual_avg}. 

\subsection{Comparison to Prior Work}

\begin{figure}[!htb]
\centering
\includegraphics[width=0.49\textwidth]{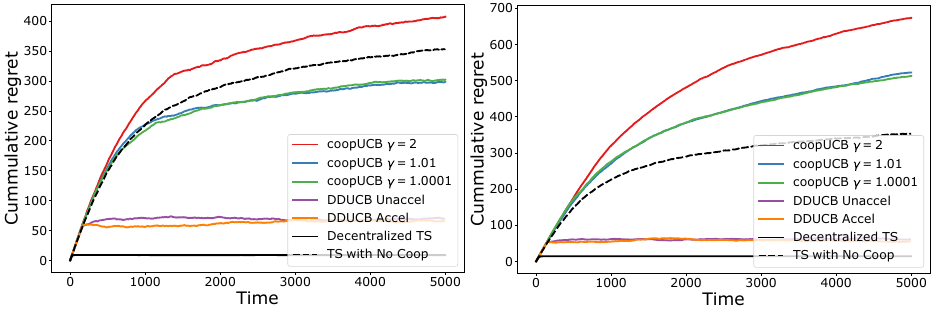}
\caption{Per-agent cumulative regret over time for $100$ agents connected in network with cycle topology (left) and grid topology (right). The agents are playing a MAB with $17$ Gaussian arms with means $\{0.5, 0.1, \ldots, 0.1\}$ and variance $\sigma^2 = 1$.}
\label{fig:dducb_langdren_decTS}
\end{figure}

We compare the performance of decentralized Thompson sampling (Algorithm~\ref{alg:decent_ts}) with coop-UCB~\cite{landgren_cdc16}, DDUCB~\cite{nips2019} and each agent implementing Thompson Sampling in isolation with no cooperation. We design an experiment similar to the one considered by Mart{\'{\i}}nez{-}Rubio et~al.~\cite{nips2019}, where each agent is playing a MAB with Gaussian rewards with variance $1$. We compare the algorithms over a cycle graph and square grid graph with 100 agents. There are $16$ arms with mean $0.1$ and one arm with mean $0.5$. All algorithms were averaged over $1000$ runs. Figure~\ref{fig:dducb_langdren_decTS} compares executions of decentralized Thompson Sampling with coop-UCB using different exploration parameters and DDUCB\footnote{We used the DDUCB implementation provided by the authors on \href{https://github.com/damaru2/decentralized-bandits}{Github}.} with the fixed exploration parameter $\eta = 2$. We observe that decentralized Thompson Sampling outperforms DDUCB and every execution of coop-UCB. Simulations of coop-UCB with different values for exploration parameter $\gamma$ show a rising slope which indicates that it has not learned the best arm. Also, note that on a square grid coop-UCB incurs a greater regret than agents implementing Thompson Sampling in isolation. This highlights the effectiveness of Bayesian algorithms such as Thompson Sampling. This also serves as an important motivation for extending Bayesian MAB algorithms and their benefits to the decentralized setup.  

\begin{figure}[!htb]
\centering
\includegraphics[width=0.5\textwidth]{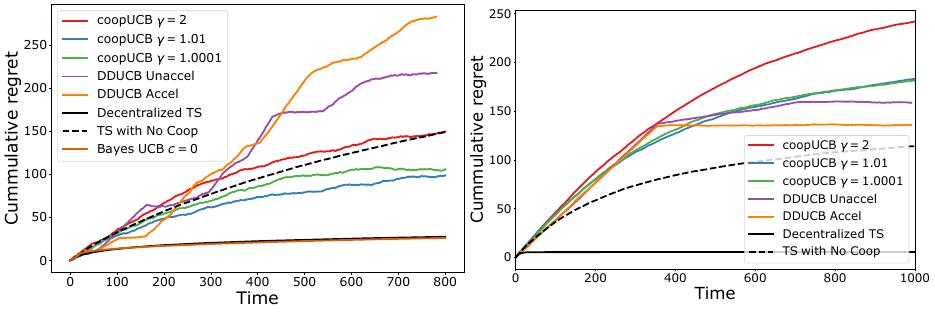}
\caption{Per-agent cumulative regret over time for a network with cycle topology with $20$ agents with $20$ Gaussian arms (left) and $200$ agents with $10$ Gaussian arms (right). Mean of the Gaussian arms were chosen uniformly randomly from $[0,1]$ and variance $\sigma^2 = 1$.}
\label{fig:dducb_langdren_decTS_bayesucb}
\end{figure}

Figure~\ref{fig:dducb_langdren_decTS_bayesucb}(left) compares executions of decentralized Thompson Sampling (Algorithm~\ref{alg:decent_ts}) and decentralized Bayes-UCB (Algorithm~\ref{alg:decent_bayesUCB}) with DDUCB and coop-UCB for $20$ agents connected through a cycle graph. Each agent is playing a MAB with $20$ Gaussian arms with variance $1$ whose means were sampled uniformly from $[0,1]$. We observe that both decentralized Bayes-UCB and decentralized Thompson Sampling outperform all executions of DDUCB and coop-UCB. 
In the figure we observe that DDUCB, decentralized Bayes-UCB, and decentralized Thompson Sampling learn the best arm after a few time steps 
and the regret curve that is observed afterwards shows an almost horizontal behavior. 
Benefits of our approach that can be inferred from this experiment: (1) our proposed decentralized Bayesian algorithms achieves the horizontal behaviour much sooner, (2) provides almost $10$x reduction in the per-agent regret. 

\subsection{Effect of Network Topology}

\begin{figure}[!htb]
\centering
\includegraphics[width=0.33\textwidth]{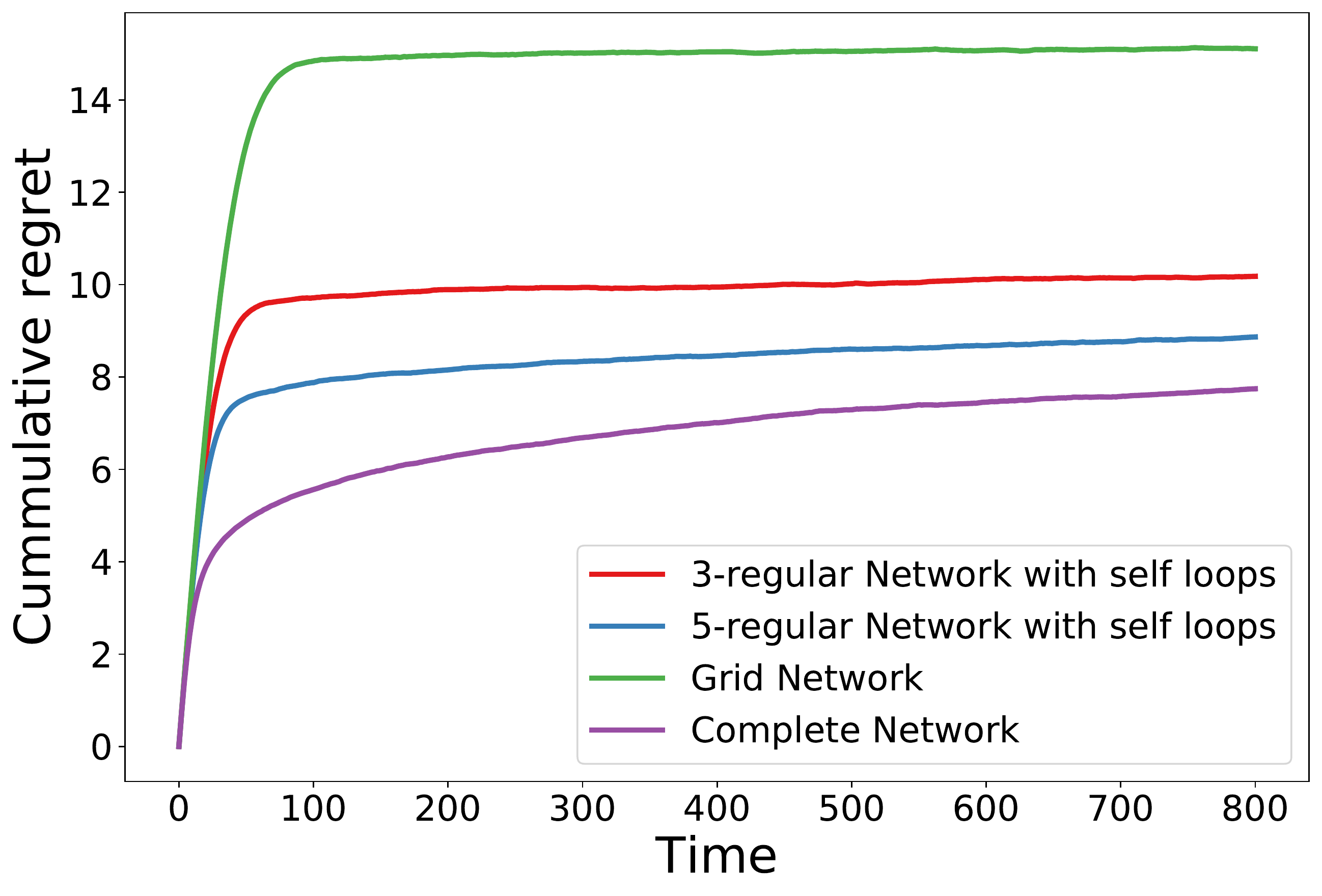}
\caption{Per-agent cumulative regret over time for $64$ agents with $17$ Bernoulli arms with mean $\{0.5, 0.1, \ldots, 0.1\}$ for varying network topology.}
\label{fig:network_topology}
\end{figure}

We consider $4$ different network topologies with $64$ agents playing a MAB with Bernoulli rewards with $17$ arms and mean values $\{0.5, 0.1, \ldots, 0.1\}$. In a network with complete graph, every agent is connected to all other agents in the network hence the spectral gap is $1$. Next, we consider $k$-regular graphs with self loops for $k=3$ and $k=5$, where every agent is connected to $k$ neighbors to the left and right and to itself. Hence, $5$-regular graph is more dense than $3$-regular. Finally we consider a two-dimensional grid with $4$-connectivity which has least connectivity among all the graphs so far. We choose $W$ according to~\cite{duchi_dual_avg}. Figure~\ref{fig:network_topology} shows that for a fixed number of agents, the per-agent regret incurred by decentralized Thompson Sampling decreases as the connectivity increases. This is in agreement with our analytical predictions in Theorem~\ref{thm:regret_decTS}, which implies that asymptotic regret is inversely proportional to the number of agents and rate of convergence is inversely proportional to spectral gap. 

\subsection{Effect of Number of Agents}

\begin{figure}[!htb]
\centering
\includegraphics[width=0.5\textwidth]{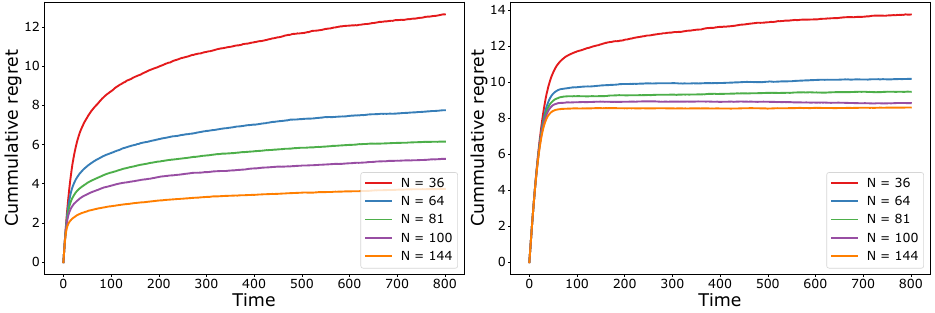}
\caption{Per-agent cumulative regret over time for agents connected in complete graph (left) and cycle graph (right) as the number of agents is varied. Each agent is playing MAB with $17$ Bernoulli arms with mean $\{0.5, 0.1, ...., 0.1\}$.}
\label{fig:network_size}
\end{figure}

We implement decentralized Thompson Sampling over a complete network and a cycle network with different number of agents. We vary the number of agents as $\{36, 64, 81, 100, 144\}$. Figure~\ref{fig:network_size} shows that the per-agent regret incurred by decentralized Thompson Sampling decreases as the number of agents increases over both complete network and cycle network.   

\subsection{Time-Varying Networks}

\begin{figure}[!htb]
\centering
\includegraphics[width=0.5\textwidth]{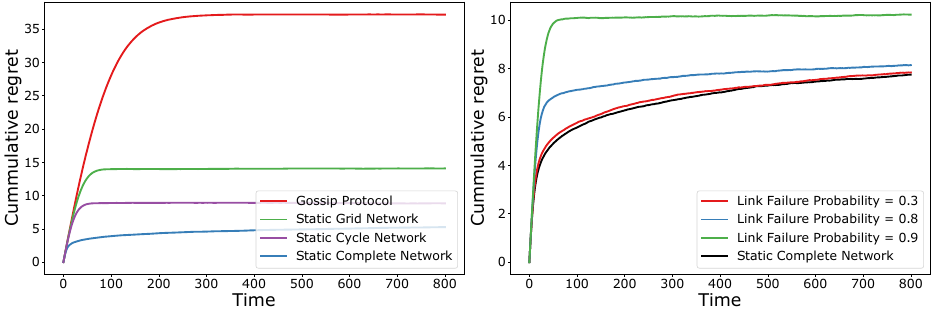}
\caption{Per-agent cumulative regret over time for $64$ agents communicating via a gossip protocol(left) and over a complete network where each communication link has a fixed probability of failing (right). Each agent plays a MAB with $17$ Gaussian arms with means $\{0.5, 0.1, \ldots, 0.1\}$ and variance $\sigma^2 = 1$.}
\label{fig:time_varying}
\end{figure}

We implement decentralized Thompson Sampling over time varying networks with $64$ agents where agent plays a MAB with $17$ Gaussian arms with means $\{0.5, 0.1, \ldots, 0.1\}$ and variance $\sigma^2 = 1$. First we consider a gossip protocol where any pair of agents $i$ and $j$ are selected uniformly at random among $64$ agents. Then, the random communication matrix is given as
\begin{align}
\label{eq:gossip}
W(t) = I - \frac{1}{2}(e_i - e_j)(e_i - e_j)^T,
\end{align}
where $I$ denotes an identity matrix and $e_i$ denotes the vector with a $1$ in the $i$th coordinate and $0$'s elsewhere. Figure~\ref{fig:time_varying}(left) shows that decentralized Thompson Sampling incurs logarithmic regret via gossip protocol. Furthermore, as predicted by the theory since connectivity of agents is significantly less when interacting via gossip protocol, i.e, communication matrix in equation~\eqref{eq:gossip}, the per-agent regret incurred is higher than that of static networks. Next, we implement decentralized Thompson Sampling over a complete network where each communication link has a fixed probability of failing. Communication matrix is chosen according to Section~7.3 in~\cite{duchi_dual_avg}. As the probability of link-failure increases the connectivity of the graph reduces.  Figure~\ref{fig:time_varying}(right) shows that decentralized Thompson Sampling incurs logarithmic regret under random link-failures where failure probability varies as $\{0.3, 0.8, 0.9\}$. It is interesting to note that for small probability of link-failure decentralized Thompson Sampling incurs same per-regret agent as a static complete network. 

\section{Conclusion and Future Work}
We have studied the problem of collaborative learning when there is a network of $N$ agents playing the same instance of the MAB problem. We design a framework to extend single-agent Bayesian MAB algorithms to the decentralized MAB problem. We use this to design a decentralized Thompson Sampling algorithm and a decentralized Bayes-UCB algorithm. For Bernoulli rewards, we provide an asymptotically optimal problem-dependent upper bound for the per-agent regret incurred for the proposed decentralized Thompson Sampling algorithm in Theorem~\ref{thm:regret_decTS}. Furthermore, we show that decentralized Thompson Sampling achieves asymptotically optimal per-agent regret that matches the performance of a centralized agent with full cooperation. Through simulations, we show that both decentralized Thompson Sampling algorithm and decentralized Bayes-UCB algorithm outperform current state-of-the-art algorithms derived from UCB.   

In decentralized settings, it is important to understand the effect of the network in terms of the number of agents and the graph structure. Currently our analytical upper bound has the right dependence on the number of agents and spectral gap. However, our simulations suggests the upper bound can be significantly tightened. We also note that our analysis provides tight upper bounds for two more important terms in equation~\eqref{eq:cond_good_events}. Improving the analysis to provide tighter regret bound in terms of network structure is essential future work. 

This work is a first step towards analyzing Bayesian algorithms for decentralized setups and motivates several future directions. Extending the analysis to a larger class of bandits or to time-varying graphs that capture practical communication constraints is an important area of future work.  Increasing the proposed algorithm's practical applicability to decentralized MAB problems with complex rewards distributions using approximate posterior update techniques such as variational inference is an important area of future work. 

\bibliographystyle{IEEEtran}
\bibliography{decentralized_bandits}

\appendix

\subsection{Preliminaries}
\label{}

\begin{itemize}
\item
Recall that $\mu_1 > \mu_2 \geq \ldots \geq \mu_K$ and $\Delta_k = \mu_1-\mu_k $. 

\item
Define history at each agent $i \in [N]$ as
\begin{align}
\mc{H}^{(i)}_{t} 
:= 
\left\{(A^{(i)}_{\tau}, Y^{(i)}_{\tau}), \tilde{Q}^{(j)}_{k,\tau}, k \in [K], j \in \mc{N}(i), \tau \in[t]
\right\}.
\end{align}

\item
For each arm $k \in [K]$, fix  thresholds $x_k$ and $y_k$ such that $\mu_k < x_k < y_k < \mu_1$. We define empirical mean estimate at each agent $i$ for action $k$ as
\begin{align}
	\hat{\mu}^{(i)}_k(t) := \frac{\alpha^{(i)}_k(t)-1}{\alpha^{(i)}_k(t) + \beta^{(i)}_k(t)-2}.
\end{align}
Now, define events
\begin{align}
	E^{(i)}_{k}(t) := \left\{\hat{\mu}^{(i)}_{k}(t)\leq x_k \right\},
\end{align}
and 
\begin{align}
	\widetilde{E}^{(i)}_{k}(t) 
	:= \left\{\theta^{(i)}_{k}(t)\leq y_k \right\}.
\end{align}

\item
Define the cumulative distribution function used for arm $k \in [K]$ at time $t$ as 
\begin{align}
	F^{(i)}_{k, t}(x) 
	:= Q^{(i)}_{k,t}\left(\theta^{(i)}_{k}(t) \leq x \right),
\end{align}
and
\begin{align}
	G^{(i)}_{k,t} := 1 - F^{(i)}_{k,t}(y_k) 
	= Q^{(i)}_{k,t}\left(\overline{\widetilde{E}^{(i)}_{k}(t)} \right).
\end{align}

\item
Define 
\begin{align}
\tilde{N} 
:= 
\frac{4N\log N}{1- \sigma_2(W)}.
\end{align}

\end{itemize}

\subsection{Main Proof}
\label{sec:main_proof}
Recall the per agent regret 
\begin{align}
R(T)
& = \frac{1}{N} \sum_{k=2}^K\Delta_k \sum_{i=1}^N\expe[n_k^{(i)}(T)].
\end{align}
Consider
\begin{align}
    &\sum_{i=1}^N\expe[n^{(i)}_k(T)]
    \\
    &= \sum_{i=1}^N\expe\left[\sum_{t=1}^T \indicate{A^{(i)}_t = k}\right]
    \\
    &= \sum_{i=1}^N\expe\left[\sum_{t=1}^T \indicate{A^{(i)}_t = k, E^{(i)}_{k}(t), \widetilde{E}^{(i)}_{k}(t)}\right] 
    \\
    &+ \sum_{i=1}^N\expe\left[\sum_{t=1}^T\indicate{A^{(i)}_t = k, E^{(i)}_{k}(t), \overline{\widetilde{E}^{(i)}_{k}(t)}}\right] \label{eq:MA_first_sec_term}
    \\
    &
    + \sum_{i=1}^N\expe\left[\sum_{t=1}^T\indicate{A^{(i)}_t = k, \overline{E^{(i)}_{k}(t)}}\right]. \label{eq:MA_third_term}
\end{align}

We next state the three main lemma that we will use in our proof. 
\begin{lemma}
\label{lemma:second_term}
Let $\eta = N$. The second term in equation~\eqref{eq:MA_first_sec_term} can be bounded as
\begin{align}
&\sum_{i=1}^N\expe\left[\sum_{t=1}^T\indicate{A^{(i)}_t = k, E^{(i)}_{k}(t), \overline{\widetilde{E}^{(i)}_{k}(t)}}\right]
\\
&\leq 
\frac{N\log N}{1-\sigma_2(W)} + \frac{\log NT}{d(x_k , y_k)} + 1.
\end{align}
\end{lemma}

\begin{lemma}
\label{lemma:third_term}
Let $\eta = N$ and $\epsilon' > 0$. The third term in equation~\eqref{eq:MA_third_term} can be bounded as
\begin{align}
&\sum_{i=1}^N\expe\left[\sum_{t=1}^T\indicate{A^{(i)}_t = k, \overline{E^{(i)}_{k}(t)}}\right]
\\
&\leq 
\frac{N\log N}{1-\sigma_2(W)}\left(1+\frac{1}{\epsilon'}\right)
+ \frac{N}{d(\frac{x_k-\epsilon'}{1+\epsilon'}, \mu_k)}.
\end{align}
\end{lemma}

\begin{lemma}
\label{lemma:first_term}
The first term in equation~\eqref{eq:MA_first_sec_term} can be bounded as
\begin{align}
&\sum_{i=1}^N\expe\left[\sum_{t=1}^T \indicate{A^{(i)}_t = k, E^{(i)}_{k}(t), \widetilde{E}^{(i)}_{k}(t)}\right] 
\\
&\leq 
\sum_{i=1}^N\expe\left[\sum_{t=1}^T\left(\frac{1}{G^{(i)}_{1,t}}-1\right)\indicate{A^{(i)}_t = 1}\right].
\end{align}
\end{lemma}

\begin{lemma}
\label{lemma:upper_bound_term1}
Let $\eta = N$, then we have 
\begin{align}
&\sum_{i=1}^N\expe\left[\sum_{t=1}^T\left(\frac{1}{G^{(i)}_{1,t}}-1\right)\indicate{A^{(i)}_t = 1}\right]
\\
&\leq 
\frac{1}{(1-y_k)^{\tilde{N}+1}}\frac{N}{(d(y_k,\mu_1))^{\tilde{N}+1}} 
+ 
\frac{N}{(y_k-\mu_1)^{\tilde{N}+1}}.
\end{align}
\end{lemma}

For some $0 < \epsilon \leq 1$, we set $x_k \in ((1+\epsilon')\mu_k + \epsilon', \mu_1)$ such that $d(x_k , \mu_1) = \frac{d(\mu_k , \mu_1)}{(1+\epsilon)}$ and set $y_k \in (x_k, \mu_1)$ such that $d(x_k , y_k) = \frac{d(x_k, \mu_1)}{(1+\epsilon)}  = \frac{d(\mu_k, \mu_1)}{(1+\epsilon)^2}$\footnote{Note that it is possible to choose $\epsilon'$ small enough such that the above equalities hold. }. This implies 
\begin{align}
\frac{1}{d(x_k, y_k)} = \frac{(1+\epsilon)^2}{d(\mu_k, \mu_1)}.
\end{align}
Furthermore, the equality $d(x_k , \mu_1) = \frac{d(\mu_k , \mu_1)}{(1+\epsilon)}$ implies
\begin{align}
x_k-\mu_k \geq \frac{\epsilon}{1+\epsilon} \frac{d(\mu_k, \mu_1)}{\log \frac{\mu_1(1-\mu_k)}{\mu_i(1-\mu_1)}} > \frac{\epsilon}{2}.
\end{align}
Using this we obtain
\begin{align}
\frac{1}{d(\frac{x_k-\epsilon'}{1+\epsilon'}, \mu_k)}
&\leq 
\frac{2(1+\epsilon')^2}{(x_k -\epsilon' - \mu_k(1+\epsilon'))^2}
\\
&\leq 
\frac{8}{(x_k - \mu_k - 2\epsilon')^2}.
\end{align}
Choose $\epsilon' = \frac{\epsilon}{8}$, then we have
\begin{align}
\frac{1}{d(\frac{x_k-\epsilon'}{1+\epsilon'}, \mu_k)}
\leq 
O\left(\frac{1}{\epsilon^2}\right).
\end{align}
Similarly, we also have
\begin{align}
	\frac{1}{d(y_k, \mu_1)}
	&\leq 
	\frac{1}{(d(x_k, \mu_1)-d(x_k, y_k))^2}
	\\
	&\leq 
	\frac{(1+\epsilon)^2}{\epsilon d(\mu_k , \mu_1)} 
	= O\left(\frac{1}{\epsilon} \right).
\end{align}
Using the fact that $\frac{\min\{\mu_1, 1-\mu_1\}}{2(y_k-\mu_1)^2} \leq \frac{1}{d(y_k, \mu_1)^2}$ along with Lemma~\ref{lemma:first_term}, we have
\begin{align}
\sum_{i=1}^N
\expe\left[\sum_{t=1}^T \indicate{A^{(i)}_t = k, E^{(i)}_{k}(t), \widetilde{E}^{(i)}_{k}(t)}\right]
&\leq 
O\left(\frac{1}{\epsilon^{\tilde{N}}}\right).
\end{align}
Therefore, the cumulative regret can be upper bounded as 
\begin{align}
R(T)
&\leq
\sum_{k=2}^K \Delta_k(1+\epsilon)^2\frac{\log NT}{N d(\mu_k , \mu_1)}
\\
&+
\frac{3\left(1+\frac{8}{\epsilon}\right)\log N}{1-\sigma_2(W)}\sum_{k=2}^K \Delta_k
+ O\left(\frac{1}{\epsilon^{\tilde{N}}} \right)
\end{align}
Asymptotically, the regret scales logarithmically with time horizon $T$ at rate upper bounded by
\begin{align}
\lim_{T \to \infty} \frac{R(T)}{\log T} 
\leq \sum_{k=2}^K\frac{\Delta_k}{Nd(\mu_k, \mu_1)}.
\end{align}

\subsection{Proof of Main Lemmata}
\label{sec:main_lemmata}

\subsubsection{Proof of Lemma~\ref{lemma:second_term}}

Set $\eta = N$. Let
\begin{align}
	\tilde{n}^{(i)}_k(t) = N \sum_{\tau = 1}^{t-1}\sum_{j=1}^N W^{t-\tau}_{ij}\indicate{A^{(j)}_{\tau} = k},
\end{align}
and 
let $\hat{n}_k(t):= n_k(t)-\tilde{N}$, then 
Lemma~\ref{lemma:alpha_beta_bounds} implies 
\begin{align}
\tilde{n}^{(i)}_k(t) \geq \hat{n}_k(t).
\end{align}
For $t$ such that $n_k(t) > \tilde{N}$, we have
\begin{align}
&\P\left(\theta^{(i)}_k(t) > y_k , E^{(i)}_k(t) \mid \mc{H}^{(i)}_{t-1}\right)
\\
& \leq 
\P\left(\theta^{(i)}_k(t) > y_k \mid E^{(i)}_k(t), \mc{H}^{(i)}_{t-1}\right)
\\
& = \P\left(\text{Beta}\left(\alpha^{(i)}_k(t), \beta^{(i)}_k(t)\right)> y_k \mid E^{(i)}_k(t) , \mc{H}^{(i)}_{t-1}\right)
\\
& = \P\left(\text{Beta}\left( \hat{\mu}^{(i)}_k(t)\tilde{n}^{(i)}_k(t)+ 1, \left(1-\hat{\mu}^{(i)}_k(t)\right) \tilde{n}^{(i)}_k(t)  + 1 \right)\right. 
\\
& \quad \quad \left. > y_k \mid E^{(i)}_k(t) , \mc{H}^{(i)}_{t-1}\right)
\\
& \overset{(a)} \leq \P\left(\text{Beta}\left( x_k\tilde{n}^{(i)}_k(t) + 1, (1-x_k) \tilde{n}^{(i)}_k(t)  + 1 \right)
> y_k \mid \mc{H}^{(i)}_{t-1}\right)
\\
& = 1- F^{\text{Beta}}_{x_k\tilde{n}^{(i)}_k(t) + 1, (1-x_k)\tilde{n}^{(i)}_k(t) + 1}(y_k)
\\
&\overset{(b)}\leq 1- F^{\text{Beta}}_{x_k\hat{n}_k(t) + 1, (1-x_k)\hat{n}_k(t) + 1}(y_k) 
\\
&\overset{(c)} = F^{\text{Binom}}_{\hat{n}_k(t)+1, y_k}(x_k \hat{n}_k(t))
\\
& \leq F^{\text{Binom}}_{\hat{n}_k(t), y_k}(x_k \hat{n}_k(t))
\\
& \overset{(d)}\leq e^{-\hat{n}_k(t) d(x_k , y_k)},
\end{align}
where $(a)$ follows from properties of beta distribution, $(b)$ follows from Fact~\ref{fact:beta_cdf2}, and $(c)$ follows from Fact~\ref{fact:beta_binom}, and $(d)$ follows from Fact~\ref{fact:chernoff}.

Define 
\begin{align}
L_k(T) := \frac{\log NT}{d(x_k , y_k)} + \tilde{N}.
\end{align}
For all $t$ such that $n_k(t) > L_k(T)$, we have 
\begin{align}
\P\left(\theta^{(i)}_k(t) > y_k, E^{(i)}_k(t) \mid \mc{H}^{(i)}_{t-1}\right) 
\leq 
\frac{1}{NT}.
\end{align}
Let $\tau_k$ be the largest time step $t$ for which $n_k(t) < L_k(T)$. Then, we have
\begin{align}
&\expe\left[\sum_{t=1}^T\indicate{A^{(i)}_t = k, E^{(i)}_{k}(t), \overline{\widetilde{E}^{(i)}_{k}(t)}}\right]
\\
& = 
\expe\left[\sum_{t = 1}^T \P\left(A^{(i)}_t = k, E^{(i)}_k(t), \overline{\widetilde{E}^{(i)}_{k}(t)} \mid \mc{H}^{(i)}_{t-1} \right)\right]
\\
&\leq 
\expe\left[\sum_{t = 1}^T \P\left(A^{(i)}_t = k,  \overline{\widetilde{E}^{(i)}_{k}(t)} \mid E^{(i)}_k(t), \mc{H}^{(i)}_{t-1} \right)\right]
\\
& = \expe\left[\sum_{t = 1}^{\tau_k} \P\left(A^{(i)}_t = k,  \overline{\widetilde{E}^{(i)}_{k}(t)} \mid E^{(i)}_k(t), \mc{H}^{(i)}_{t-1} \right)\right.
\\
& \quad \quad  \left.+ \sum_{t = \tau_k+1}^T \P\left(A^{(i)}_t = k,  \overline{\widetilde{E}^{(i)}_{k}(t)} \mid E^{(i)}_k(t), \mc{H}^{(i)}_{t-1} \right)\right]
\\
& \leq \expe\left[\sum_{t = 1}^{\tau_k} \P\left(A^{(i)}_t = k,  \overline{\widetilde{E}^{(i)}_{k}(t)} \mid E^{(i)}_k(t), \mc{H}^{(i)}_{t-1} \right)\right.
\\
& \quad \quad \left.+ \sum_{t = \tau_k+1}^T \frac{1}{NT}\right]
\\
& \leq \expe\left[\sum_{t = 1}^{\tau_k} \P\left(A^{(i)}_t = k,  \overline{\widetilde{E}^{(i)}_{k}(t)} \mid E^{(i)}_k(t), \mc{H}^{(i)}_{t-1} \right)\right] + \frac{1}{N}
\\
& \leq \expe\left[\sum_{t = 1}^{\tau_k} \indicate{A^{(i)}_t = k}\right] + \frac{1}{N}.
\end{align}
Summing over all agents in the network, the above equation becomes
\begin{align}
&\sum_{i=1}^N \expe\left[\sum_{t=1}^T\indicate{A^{(i)}_t = k, E^{(i)}_{k}(t), \overline{\widetilde{E}^{(i)}_{k}(t)}}\right]
\\
& \leq
\expe\left[\sum_{t = 1}^{\tau_k}\sum_{i=1}^N \indicate{A^{(i)}_t = k}\right] + \sum_{i=1}^N\frac{1}{N}
\\
& \leq L_k(T) + 1. 
\end{align} 

\subsubsection{Proof of Lemma~\ref{lemma:third_term}}

From the definition of $\hat{\mu}^{(i)}_k(t)$ and Lemma~\ref{lemma:posterior_closed_form}, we have 
\begin{align}
\hat{\mu}^{(i)}_k(t) 
& = 
\frac{\alpha^{(i)}_k(t)-1}{\alpha^{(i)}_k(t)+\beta^{(i)}_k(t)-2}
\\
& = 
\frac{\sum_{\tau = 1}^{t-1}\sum_{j=1}^{N}W^{t-\tau}_{ij}Y_{\tau,k}^{(j)}\indicate{A^{(j)}_{\tau} = k}}{\sum_{\tau = 1}^{t-1}\sum_{j=1}^{N}W^{t-\tau}_{ij}\indicate{A^{(j)}_{\tau} = k}}.
\end{align}
Then, using Lemma~\ref{lemma:alpha_beta_bounds}, for $n_k(t) > \tilde{N}$ we can upper the empirical mean as follows
\begin{align}
\hat{\mu}^{(i)}_k(t) 
& \leq
\frac{S_k(t) + \tilde{N}}{n_k(t) - \tilde{N}}
\\
& = \frac{S_k(t)}{n_k(t)}\cdot\frac{n_k(t)}{n_k(t) - \tilde{N}}
+ \frac{\tilde{N}}{n_k(t) - \tilde{N}}.
\end{align}

 

\begin{figure}[!htb]
\centering
\includegraphics[width=0.5\textwidth]{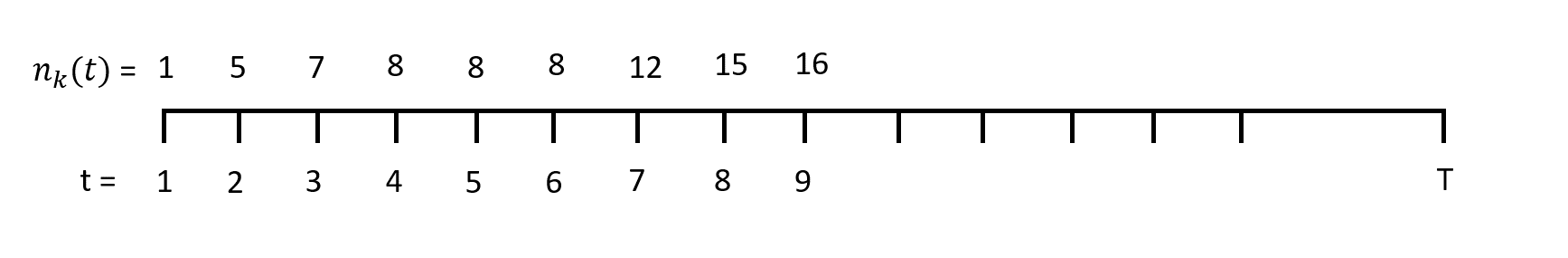}
\caption{A sample path illustrating the number of times arm $k$ was pulled at each time instant when $N = 4$.}
\label{fig:lemma_3}
\end{figure}

Recall that $n_k(t)$ denotes the total number of times arm $k$ was pulled in the network up to time $t$. Let $\tau_{k, n}$ denote the smallest $t$ when arm $k$ was played for the $n$-th time in the network, i.e., define $\tau_{k, n} = \min\{t: n_k(t) = n\}$ for any integer $n$. Note that if $\tau_{k,n} = t-1$ for some $t$, and at time $t$, suppose $\ell \leq N$ agents play arm $k$ at the same time then we define $\tau_{k, n+1} = t$, $\tau_{k, n+2} = t$, and so on up to $\tau_{k, n+\ell} = t$. The sequence $\{\tau_{k,n}\}_{n =1}^{NT}$ counts some time steps more than once, hence, we choose a sub-sequence $\{n_j\}_{j \geq 1}$, which denotes the change points in the sequence $\{\tau_{k,n}\}_{n \geq 1}$.  For instance, Figure~\ref{fig:lemma_3} shows a sample path of the number times action $k$ was played at each time instant in a network of 4 agents. We see that $\tau_{k,1} = 1$, $\tau_{k,2} = \tau_{k,3} = \tau_{k,4} = \tau_{k,5} = 2$, and $\tau_{k,6} = \tau_{k,7} = 3$, etc. The change points sub-sequence for this case is given by $\tau_{k,1}, \tau_{k, 5}, \tau_{k, 7}, \tau_{k, 8}, \tau_{k,12}$ and so on.

Then, we have
\begin{align}
    &\sum_{i=1}^N \sum_{t = 1}^T \P\left( A^{(i)}_t = k, \overline{E^{(i)}_k(t)}\right)
    \\
    & = \expe\left[ \sum_{i=1}^N\sum_{t = 1}^{T}\indicate{A^{(i)}_t = k}\indicate{\overline{E^{(i)}_k(t)}}\right]
    \\
    & = \expe\left[\sum_{j = 1}^{J_T}\sum_{t = \tau_{k, n_j+1}}^{\tau_{k,n_{j+1}}} \sum_{i=1}^N\indicate{A^{(i)}_t = k}\indicate{\overline{E^{(i)}_k(t)}}\right].
\end{align}

Let $n_j > \tilde{N}$. For $t \in \{\tau_{k,n_{j}+1}, \tau_{k,n_{j}+2}, \ldots, \tau_{k,n_{j+1}}\}$, the total number of times arm $k$ was pulled across the network remains $n_j$. Hence, we have 
\begin{align}
\left\{\hat{\mu}_k^{(i)}(t) > x_k\right\}
&\subset E_k(n_j),
\end{align}
where define 
\begin{align}
E_k(j) := 
\left\{
\frac{\sum_{\tau = 1}^{j} X_{k, \tau}}{j}\cdot \frac{j}{j-\tilde{N}}
 + \frac{\tilde{N}}{j - \tilde{N}}
> x_k
\right\},
\end{align}
for $j > \tilde{N}$.

Now consider
\begin{align}
    & \sum_{i=1}^N \sum_{t = 1}^T \P\left( A^{(i)}_t = k, \overline{E^{(i)}_k(t)}\right)
    \\
    & = \expe\left[\sum_{j = 1}^{J_T}\sum_{t = \tau_{k, n_j+1}}^{\tau_{k,n_{j+1}}} \sum_{i=1}^N\indicate{A^{(i)}_t = k}\indicate{\overline{E^{(i)}_k(t)}}\right]
    \\
    & \leq \expe\left[\sum_{j = 1}^{\tilde{N}}\sum_{t = \tau_{k, n_j+1}}^{\tau_{k,n_{j+1}}} \sum_{i=1}^N\indicate{A^{(i)}_t = k}\right]
    \\
    & \quad + \expe\left[\sum_{j = \tilde{N}+1}^{J_T}\sum_{t = \tau_{k, n_j+1}}^{\tau_{k,n_{j+1}}} \sum_{i=1}^N\indicate{A^{(i)}_t = k}\indicate{\overline{E^{(i)}_k(t)}}\right]
    \\
    & \leq \tilde{N} + 
    \expe\left[\sum_{j = \tilde{N}+ 1}^{J_T}\sum_{t = \tau_{k, n_j+1}}^{\tau_{k,n_{j+1}}} \sum_{i=1}^N\indicate{A^{(i)}_t = k}\indicate{E_k(n_j)}\right]
    \\
    & = \tilde{N} + 
    \expe\left[\sum_{j = \tilde{N}+ 1}^{J_T}\indicate{E_k(n_j)}\sum_{t = \tau_{k, n_j+1}}^{\tau_{k,n_{j+1}}} \sum_{i=1}^N\indicate{A^{(i)}_t = k}\right]
    \\
    &\overset{(a)} \leq \tilde{N} + 
    N\expe\left[\sum_{j = \tilde{N} + 1}^{J_T}\indicate{E_k(n_j)}\right]
    \\
    & \leq 
    \tilde
    {N} +  N\expe\left[\sum_{j = \tilde{N}+1}^{NT}\indicate{E_k(j)}\right],
\end{align}
where $(a)$ follows from the fact the sub-sequence $\{n_j\}$ detects a change whenever one or more agents pull the arm $k$ and at any time at most $N$ agents can pull the same arm.

Fix some $\epsilon' > 0$, then for $j > \tilde{N}\left(1+\frac{1}{\epsilon'}\right)$, we have $\frac{\tilde{N}}{j-\tilde{N}} < \epsilon'$ and $\frac{j}{j-\tilde{N}} < 1 + \epsilon'$. This implies 
\begin{align}
E_k(j) \subset 
\left\{
\frac{\sum_{\tau = 1}^{j} X_{k, \tau}}{j}
> \frac{x_k-\epsilon'}{1+\epsilon'}
\right\}.
\end{align}  
Hence, we have
\begin{align}
    &\sum_{i=1}^N \sum_{t = 1}^T \P\left( A^{(i)}_t = k, \overline{E^{(i)}_k(t)}\right)
    \\
    &\leq 
    \tilde{N}\left(1+\frac{1}{\epsilon'}\right) + N\expe\left[\sum_{j=\tilde{N}(1+1/\epsilon')}^{NT}\indicate{E_k(j)}\right]
    \\
    & \leq  
    \tilde{N}\left(1+\frac{1}{\epsilon'}\right) + N\sum_{j=\tilde{N}(1+1/\epsilon')}^{\infty}\P\left(\frac{\sum_{\tau = 1}^{j} X_{k, \tau}}{j}
    > \frac{x_k-\epsilon'}{1+\epsilon'}
    \right)
    \\
    & \overset{(a)}\leq  
    \tilde{N}\left(1+\frac{1}{\epsilon'}\right)
    + N\sum_{j=0}^{\infty} e^{-jd(\frac{x_k-\epsilon'}{1+\epsilon'}, \mu_k)}
    \\
    & \leq 
    \tilde{N}\left(1+\frac{1}{\epsilon'}\right)
    + \frac{N}{d(\frac{x_k-\epsilon'}{1+\epsilon'}, \mu_k)}
\end{align}
where $(a)$ follows from Chernoff bound given in Fact~\ref{fact:chernoff}.

\subsubsection{Proof of Lemma~\ref{lemma:first_term}}

Let $A^{\prime (i)}_t = \argmax_{k \neq 1} \theta^{(i)}_{k}(t)$. Suppose that the filtration is such that $E^{(i)}_k(t)$ is true, otherwise the probability on the left hand side is 0 and the inequality is trivially true. Then
\begin{align}
	&\P(A^{(i)}_t = 1, \widetilde{E}^{(i)}_k(t) \mid \mc{H}^{(i)}_{t-1})
	\\
	& \geq \P(A^{\prime (i)}_t = k, \widetilde{E}^{(i)}_k(t), \{\theta^{(i)}_1(t) > y_k \} \mid \mc{H}^{(i)}_{t-1})
	\\
	&\overset{(a)} = \P(\theta^{(i)}_1(t) >y_k \mid \mc{H}^{(i)}_{t-1})\P(A^{\prime (i)}_t = k, \widetilde{E}^{(i)}_k(t)\mid \mc{H}^{(i)}_{t-1})
	\\
	& = G^{(i)}_{1,t}\P(A^{\prime (i)}_t = k, \widetilde{E}^{(i)}_k(t)\mid \mc{H}^{(i)}_{t-1}),
\end{align}
where in $(a)$ we used the fact that $\theta^{(i)}_1(t)$ is conditionally independent of $A^{\prime (i)}_t$ and $\widetilde{E}^{(i)}_{k}(t)$ given $\mc{H}^{(i)}_{t-1}$. Similarly, we have
\begin{align}
	&\P(A^{(i)}_t = k, \widetilde{E}^{(i)}_k(t) \mid \mc{H}^{(i)}_{t-1})
	\\
	& \leq \P(A^{\prime (i)}_t = k, \widetilde{E}^{(i)}_k(t), \{\theta^{(i)}_1(t) \leq y_k \} \mid \mc{H}^{(i)}_{t-1})
	\\
	&\overset{(b)} = (1-G^{(i)}_{1,t}) \P(A^{\prime (i)}_t = k, \widetilde{E}^{(i)}_k(t) \mid \mc{H}^{(i)}_{t-1}),
\end{align}
where in $(b)$ we again used the fact that $\theta^{(i)}_1(t)$ is conditionally independent of $A^{\prime (i)}_t$ and $\widetilde{E}^{(i)}_{k}(t)$ given $\mc{H}^{(i)}_{t-1}$. Hence, we have
\begin{align}
	&\P(A^{(i)}_t = k, \widetilde{E}^{(i)}_k(t) \mid \mc{H}^{(i)}_{t-1})
	\\
	& \leq 
	\left(\frac{1-G^{(i)}_{1,t}}{G^{(i)}_{1,t}}\right)\P(A^{(i)}_t = 1 , \widetilde{E}^{(i)}_{k}(t) \mid \mc{H}^{(i)}_{t-1})
	\\
	& \leq 
	\left(\frac{1}{G^{(i)}_{1,t}}-1\right)\P(A^{(i)}_t = 1 \mid \mc{H}^{(i)}_{t-1}). 
\end{align} 
The first term can be bounded as
\begin{align}
&\expe\left[\sum_{t=1}^T \indicate{A^{(i)}_t = k, E^{(i)}_{k}(t), \widetilde{E}^{(i)}_{k}(t)}\right] 
\\
& \leq 
\expe\left[\sum_{t=1}^T\left(\frac{1}{G^{(i)}_{1,t}}-1\right)\P(A^{(i)}_t = 1 \mid \mc{H}^{(i)}_{t-1})\right]
\\
& = 
\expe\left[\sum_{t=1}^T\left(\frac{1}{G^{(i)}_{1,t}}-1\right)\indicate{A^{(i)}_t = 1}\right].
\end{align}
Summing over the number of agents we have the assertion of the lemma.

\subsubsection{Proof of Lemma~\ref{lemma:upper_bound_term1}}

Note that 
\begin{align}
&\expe\left[\sum_{t=1}^T\left(\frac{1}{G^{(i)}_{1,t}}-1\right)\indicate{A^{(i)}_t = 1}\right]
\\
& = 
\expe\left[\sum_{t=1}^T\left(\frac{1}{\q^{(i)}_{1,t}(\theta^{(i)}_{1}(t) > y_k)}-1\right)\indicate{A^{(i)}_t = 1}\right].
\end{align}

Now, consider
\begin{align}
&\q^{(i)}_{1,t}\left(\theta^{(i)}_{1}(t) > y_k \right)
\\
& =
\P\left(
\text{Beta}\left(\alpha^{(i)}_{1}(t), \beta^{(i)}_1(t)\right)
> y_k
\mid \mc{H}^{(i)}_{t-1} \right)
\\
& = 1-F^{\text{Beta}}_{\alpha^{(i)}_1(t),\, \beta^{(i)}_1(t)}(y_k)
\\
&\overset{(a)} \geq 1- F^{\text{Beta}}_{S_1(t)-\tilde{N}+1,\, n_1(t)-S_1(t) + \tilde{N}+1}(y_k),
\end{align}
where $(a)$ follows from Lemma~\ref{lemma:alpha_beta_bounds} combined with properties of Beta distribution in Fact~\ref{fact:beta_cdf1}. Let $S_1'(t) = \max\{S_1(t)-\tilde{N}, 0\}$. Using this the first term can be upper bounded as
\begin{align}
&\sum_{i=1}^N
\expe\left[\sum_{t=1}^T \indicate{A^{(i)}_t = k, E^{(i)}_{k}(t), \widetilde{E}^{(i)}_{k}(t)}\right] 
\\
& \leq \expe\left[\sum_{i=1}^N \sum_{t=1}^T\left(\frac{1}{\q^{(i)}_{1,t}(\theta^{(i)}_{1}(t) > y_k)}-1\right)\indicate{A^{(i)}_t = 1}\right]
\\
& \leq
\expe\left[
\sum_{j = 1}^{J_T}\sum_{t = \tau_{1, n_j+1}}^{\tau_{1,n_{j+1}}} \sum_{i=1}^N \left( \frac{1}{1- F^{\text{Beta}}_{S'_1(t)+1,\, n_1(t)-S_1(t) + \tilde{N}+1}(y_k)}\right.\right.
\\
&\quad \quad \quad \left.\left.-1 \right)\indicate{A^{(i)}_t = 1}
\right]
\\
&\overset{(a)} =
\expe\left[
\sum_{j = 1}^{J_T}\sum_{t = \tau_{1, n_j+1}}^{\tau_{1,n_{j+1}}}  \left( \frac{1}{1- F^{\text{Beta}}_{S'_1(n_j)+1,\, n_j-S_1(n_j) + \tilde{N}+1}(y_k)}\right.\right.
\\
&\quad \quad \quad \left.\left.-1 \right)\sum_{i=1}^N \indicate{A^{(i)}_t = 1}
\right] 
\\
& \leq
\expe\left[
\sum_{j = 1}^{J_T}  \left( \frac{1}{1- F^{\text{Beta}}_{S'_1(n_j)+1,\, n_j-S_1(n_j) + \tilde{N}+1}(y_k)}-1 \right)\right.
\\
&\quad \quad \left.\sum_{t = \tau_{1, n_j+1}}^{\tau_{1,n_{j+1}}}\sum_{i=1}^N \indicate{A^{(i)}_t = 1}
\right] 
\\
& \leq
N\expe\left[
\sum_{j = 1}^{J_T}  \left( \frac{1}{1- F^{\text{Beta}}_{S'_1+1,\, n_j-S_1(n_j) + \tilde{N}+1}(y_k)}-1 \right)
\right] 
\\
& \leq
N\expe\left[
\sum_{j = 1}^{NT}  \left( \frac{1}{1- F^{\text{Beta}}_{S'_1(j)+1,\, j-S_1(j) + \tilde{N}+1}(y_k)}-1 \right)
\right],
\end{align}
where $(a)$ follows from the fact that at $n_1(t) = n_j$ for $t \in [\tau_{1, n_j}+1, \tau_{1, n_{j+1}}]$ and we abuse the notation let $S_1(n_j)$ denote the number of non-zero rewards observed in the first $n_j$ arm plays in the network. Now, we consider
\begin{align}
&\expe\left[
\frac{1}{1-F^{\text{Beta}}_{(S_1(j)-\tilde{N})^{+}+1,j-S_1(j)+\tilde{N}+1 }(y_1)}-1
\right]
\\
& \leq 
\sum_{s=0}^{j}
\frac{\binom{j}{s} \mu_1^{s} (1-\mu_1)^{j-s}}{F^{\text{Binom}}_{j+1, y_1}(s-\tilde{N})}
\\
& = \sum_{s=0}^{\lfloor y_1j \rfloor + \tilde{N}}
\frac{\binom{j}{s} \mu_1^{s} (1-\mu_1)^{j-s}}{F^{\text{Binom}}_{j+1, y_1}(s-\tilde{N})}
\\
& \quad \quad  +
\sum_{s=\lfloor y_1j \rfloor + \tilde{N} + 1}^{j}
\frac{\binom{j}{s} \mu_1^{s} (1-\mu_1)^{j-s}}{F^{\text{Binom}}_{j+1, y_1}(s-\tilde{N})}
\\
& \leq \sum_{k=0}^{\lfloor y_1j \rfloor }
\frac{\binom{j}{k+\tilde{N}} \mu_1^{k+\tilde{N}} (1-\mu_1)^{j-k-\tilde{N}}}{\binom{j+1}{k} y_1^{k} (1-y_1)^{j+1-k}}
\\
& \quad \quad  + 
\sum_{s=\lfloor y_1j \rfloor +1}^{s }
2\binom{j}{s} \mu_1^{s} (1-\mu_1)^{j-s}
\\
& \leq \frac{1}{(1-y_1)^{\tilde{N}+1}}\sum_{k=0}^{\lfloor y_1j \rfloor }
\frac{\binom{j}{k+\tilde{N}} \mu_1^{k+\tilde{N}} (1-\mu_1)^{j-k-\tilde{N}}}{\binom{j}{k} y_1^{k+\tilde{N}} (1-y_1)^{j-k-\tilde{N}}}
\\
& \quad \quad + 
\sum_{s=\lfloor y_1j \rfloor +1}^{s }
2\binom{j}{s} \mu_1^{s} (1-\mu_1)^{j-s}
\\
& 
\overset{(a)}\leq \frac{1}{(1-y_1)^{\tilde{N}+1}}\frac{j^{\tilde{N}}}{\tilde{N}!}\sum_{k=0}^{\lfloor y_1j \rfloor }
\frac{ \mu_1^{k+\tilde{N}} (1-\mu_1)^{j-k-\tilde{N}}}{ y_1^{k+\tilde{N}} (1-y_1)^{j-k-\tilde{N}}}
\\
& \quad \quad + 
\sum_{s=\lfloor y_1j \rfloor +1}^{s }
2\binom{j}{s} \mu_1^{s} (1-\mu_1)^{j-s}
\\
& \leq 
\frac{1}{(1-y_1)^{\tilde{N}+1}}\frac{j^{\tilde{N}}}{\tilde{N}!}\sum_{k=0}^{\lfloor y_1j \rfloor }
\frac{ \mu_1^{k} (1-\mu_1)^{j-k}}{ y_1^{k} (1-y_1)^{j-k}}
\\
& \quad \quad + 
\frac{4j^{\tilde{N}}}{\tilde{N}!}\sum_{s=\lfloor y_1j \rfloor +1}^{s }
\binom{j}{s} \mu_1^{s} (1-\mu_1)^{j-s}
\\
&\overset{(b)} \leq 
\frac{1}{(1-y_1)^{\tilde{N}+1}}\frac{j^{\tilde{N}}}{\tilde{N}!}\mu_1 e^{-jd(y_1, \mu_1)}
+ 
\frac{4j^{\tilde{N}}}{\tilde{N}!} e^{-2j(y_1-\mu_1)^2},
\end{align}
where $(a)$ follows from the fact that $\frac{\binom{j}{k+\tilde{N}}}{\binom{j}{k}} \leq \frac{j^{\tilde{N}}}{\tilde{N}!}$ and $(b)$ follows from Chernoff bound in Fact~\ref{fact:chernoff} and direct computation as shown in proof of Theorem~2 in S-6.1 in~\cite{zhu2020thompson}. Summing over over $j$ and following proof Theorem~2 in S-6.1 in~\cite{zhu2020thompson}, we have
\begin{align}
&\sum_{j=1}^{\infty}\expe\left[
\frac{1}{1-F^{\text{Beta}}_{(S_1(j)-\tilde{N})^{+}+1,j-S_1(j)+\tilde{N}+1 }(y_1)}-1
\right]
\\
&\leq 
\frac{1}{(1-y_1)^{\tilde{N}+1}}\int_0^{\infty}\frac{j^{\tilde{N}}}{\tilde{N}!}\mu_1 e^{-jd(y_1, \mu_1)}dj
\\
& \quad \quad 
+ \int_0^{\infty}\frac{j^{\tilde{N}}}{\tilde{N}!} e^{-2j(y_1-\mu_1)^2}dj
\\
&\overset{(a)} = 
\frac{1}{(1-y_1)^{\tilde{N}+1}}\frac{1}{\tilde{N}!} \frac{\tilde{N}!}{(d(y_1,\mu_1))^{\tilde{N}+1}} 
\\
& \quad \quad + 
\frac{1}{\tilde{N}!}  \frac{\tilde{N}!}{2^{\tilde{N}+1}(y_1-\mu_1)^{\tilde{N}+1}}
\\
& = 
\frac{1}{(1-y_1)^{\tilde{N}+1}}\frac{1}{(d(y_1,\mu_1))^{\tilde{N}+1}} 
\\
& \quad \quad + 
\frac{1}{2^{\tilde{N}+1}(y_1-\mu_1)^{\tilde{N}+1}}
\\
& \leq 
\frac{1}{(1-y_1)^{\tilde{N}+1}}\frac{1}{(d(y_1,\mu_1))^{\tilde{N}+1}} 
\\
& \quad \quad + 
\frac{1}{(y_1-\mu_1)^{\tilde{N}+1}}
\end{align}
where $(a)$ follows from the identity $\int_0^{\infty} x^n e^{-ax}dx = \frac{n!}{a^{n+1}}$. Summing over the number of agents we have
\begin{align}
&\sum_{i=1}^N
\expe\left[\sum_{t=1}^T \indicate{A^{(i)}_t = k, E^{(i)}_{k}(t), \widetilde{E}^{(i)}_{k}(t)}\right]
\\
&\leq 
\frac{1}{(1-y_1)^{\tilde{N}+1}}\frac{N}{(d(y_1,\mu_1))^{\tilde{N}+1}} 
+ 
\frac{N}{(y_1-\mu_1)^{\tilde{N}+1}}.
\end{align}

\subsection{Proofs for additional lemmata}
\label{sec:additional_lemmata}

\begin{lemma}
\label{lemma:posterior_closed_form}
At any given time $t$, for any agent $j \in [N]$, the local parameter $\theta^{(i)}_k(t)$ for any arm $k \in [K]$ is sampled from a posterior distribution which is beta distribution $ \mathsf{Beta}\left(\alpha^{(i)}_k(t), \beta^{(i)}_k(t)\right)$, whose parameters are given by
\begin{align}
\label{eq:alpha_closed_form}
\alpha^{(i)}_{k}(t) = \eta \sum_{\tau = 1}^{t-1}\sum_{j=1}^{N}W^{t-\tau}_{ij}Y_{\tau,k}^{(j)}\indicate{A^{(j)}_{\tau} = k} + 1,
\end{align}
and
\begin{align}
\label{eq:beta_closed_form}
\beta^{(i)}_k(t) = \eta \sum_{\tau = 1}^{t-1}\sum_{j=1}^{N}W^{t-\tau}_{ij}(1-Y_{\tau,k}^{(j)})\indicate{A^{(j)}_{\tau} = k} + 1.
\end{align}
In other words, we have
\begin{align}
Q^{(i)}_{k,t} = \mathsf{Beta}\left(\alpha^{(i)}_k(t), \beta^{(i)}_k(t) \right), \forall i \in [N], k \in [K].
\end{align}
\end{lemma}

\begin{IEEEproof}
Let $\mathsf{Beta}(\tilde{\alpha}_k^{(i)}(t), \tilde{\beta}_k^{(i)}(t))$ denote the beta posterior distribution at agent $i$ for arm $k$ before merging with the neighbors posteriors at time $t$. Let $f(\theta \mid \alpha, \beta)$ denote the pdf of $\mathsf{Beta}(\alpha, \beta)$. Let $B(\alpha, \beta) = \int_{0}^{1}u^{\alpha-1}(1-u)^{\beta-1}du$ denote the beta function. Recall that posterior merging at agent $i$ is performed as follows
\begin{align}
\label{eq:merging_beta_post}
\frac{\exp{\sum_{j=1}^NW_{ij} \log f(\theta \mid \tilde{\alpha}_k^{(j)}(t), \tilde{\beta}_k^{(j)}(t))}}{\int_{u} \exp{\sum_{j=1}^N}W_{ij} \log f(u \mid \tilde{\alpha}_k^{(j)}(t), \tilde{\beta}_k^{(j)}(t))du}.
\end{align}
Numerator of the term in~\eqref{eq:merging_beta_post} can be expanded as follows
\begin{align}
&\exp{\left( \sum_{j=1}^NW_{ij} \log f(\theta \mid \tilde{\alpha}_k^{(j)}(t), \tilde{\beta}_k^{(j)}(t)) \right)}
\\
& = \theta ^{\left(  \sum_{j=1}^N W_{ij} 
\tilde{\alpha}^{(j)}_k(t) - 1\right)} 
(1-\theta)^{ \left(  \sum_{j=1}^N W_{ij} 
\tilde{\beta}^{(j)}_k(t) - 1\right)} \times
\\
&\quad \quad \times \exp{
\left(
\sum_{j=1}^N W_{ij} \log B(\tilde{\alpha}^{(j)}_k(t), \tilde{\beta}^{(j)}_k(t))\right)}.
\label{eq:numerator_merging} 
\end{align}
Similarly, denominator of the term in~\eqref{eq:merging_beta_post} equals
\begin{align}
& \int_0^1 u^{\left(  \sum_{j=1}^N W_{ij} 
\tilde{\alpha}^{(j)}_k(t) - 1\right)}
(1-u)^{\left(  \sum_{j=1}^N W_{ij} 
\tilde{\beta}^{(j)}_k(t) - 1\right)} du \times 
\\
& \times \exp{\left(\sum_{j=1}^N W_{ij} \log B(\tilde{\alpha}^{(j)}_k(t), \tilde{\beta}^{(j)}_k(t))\right)}.
\label{eq:denominator_merging}
\end{align}

Combining equation~\eqref{eq:numerator_merging} and~\eqref{eq:denominator_merging} we have
\begin{align}
\frac{\theta ^{\left(  \sum_{j=1}^N W_{ij} 
\tilde{\alpha}^{(j)}_k(t) - 1\right)} 
(1-\theta)^{ \left(  \sum_{j=1}^N W_{ij} 
\tilde{\beta}^{(j)}_k(t) - 1\right)\log (1-\theta)}}{B\left(\sum_{j=1}^N W_{ij} 
\tilde{\alpha}^{(j)}_k(t), \sum_{j=1}^N W_{ij}\tilde{\beta}^{(j)}_k(t)\right)}.
\end{align}
Hence we have 
\begin{align}
&\alpha^{(i)}_k(t) 
= \sum_{j=1}^N W_{ij} \tilde{\alpha}^{(j)}_k(t) 
\\
&= \sum_{j=1}^N W_{ij}\alpha^{(j)}_k(t-1) + \sum_{j=1}^N \eta W_{ij} Y^{(j)}_{t,k}\indicate{A^{(j)}_{t}  = k},
\\
&\beta^{(i)}_k(t) 
= \sum_{j=1}^N W_{ij}\tilde{\beta}^{(j)}_k(t) 
\\
&= \sum_{j=1}^N W_{ij}\beta^{(j)}_k(t-1) + \sum_{j=1}^N \eta W_{ij} (1-Y^{(j)}_{t,k})\indicate{A^{(j)}_{t}  = k}.
\end{align}
Applying the above equations recursively, we obtain equations~\eqref{eq:alpha_closed_form} and~\eqref{eq:beta_closed_form}. 
\end{IEEEproof}

Lemma~\ref{lemma:beta_bernoulli_update} follows from the following lemma.

\begin{lemma}
\label{lemma:alpha_beta_bounds}
The parameters of the beta posterior distribution at every agent $i \in [N]$ can be bounded as follows
\begin{align}
\alpha^{(i)}_{k}(t)
& \leq 
\frac{\eta}{N}S_k(t) + 1 + \frac{4\eta\log N}{1- \sigma_2(W)} ,
\\
\alpha^{(i)}_{k}(t)
& \geq
\frac{\eta}{N}S_k(t) + 1 - \frac{4\eta\log N}{1- \sigma_2(W)},
\end{align}
and
\begin{align}
\beta^{(i)}_{k}(t)
& \leq 
\frac{\eta}{N}(n_k(t) - S_k(t)) + 1 + \frac{4\eta\log N}{1- \sigma_2(W)},
\\
\beta^{(i)}_{k}(t)
& \geq \frac{\eta}{N}(n_k(t) - S_k(t)) + 1 - \frac{4\eta\log N}{1- \sigma_2(W)},
\end{align}
where $S_k(t)$ denotes the sum of rewards coming from all the pulls done for arm $k$ by the entire network upto time $t$ and $n_{k}(t)$ denotes the total number of times arm $k$ was pulled across the network upto time $t$. 
\end{lemma}

\begin{IEEEproof}

Consider
\begin{align}
\alpha^{(i)}_{k}(t)
& \overset{(a)} = \eta\sum_{\tau=1}^{t-1}\sum_{j=1}^N W^{t-\tau}_{ij} Y^{(j)}_{\tau, k}\indicate{A^{(j)}_{\tau}=k} +  1
\\
& = \eta\sum_{\tau=1}^{t}\sum_{j=1}^N \left(W^{t-\tau}_{ij}-\frac{1}{N}\right) Y^{(j)}_{\tau, k}\indicate{A^{(j)}_{\tau}=k}
\\
& \quad \quad +
\frac{\eta}{N}\sum_{\tau=1}^{t-1}\sum_{j=1}^N  Y^{(j)}_{\tau, k}\indicate{A^{(j)}_{\tau}=k} 
+ 
1
\\
& \overset{(b)}\leq \eta\sum_{\tau=1}^{t}\sum_{j=1}^N \left| W^{t-\tau}_{ij}-\frac{1}{N}\right| 
\\
& \quad \quad +
\frac{\eta}{N}\sum_{\tau=1}^{t-1}\sum_{j=1}^N  Y^{(j)}_{\tau, k}\indicate{A^{(j)}_{\tau}=k} 
+ 
1
\\
& \overset{(c)}\leq 
\frac{4\eta \log N}{1-\sigma_2(W)} + \frac{\eta}{N} S_k(t) + 1
\end{align}
where $(a)$ follows from Lemma~\ref{lemma:posterior_closed_form}, $(b)$ follows from fact that rewards are upper bounded by 1, $(c)$ follows from Fact~\ref{fact:W_sum_conv}.  Rest of the proof follows similarly.
\end{IEEEproof}

\subsubsection{Proof of Lemma~\ref{lemma:gaussian_update}}
First part of the lemma follows by applying Bayes rule. For merging the posterior consider
\begin{align}
&\sum_{j=1}^{N} W_{ij} \log f(\theta \mid \tilde{\mu}^{(j)}_k(t), (\tilde{\sigma}^{(j)}_k(t))^2)
\\
&= 
- \frac{1}{2}\sum_{j=1}^N W_{ij}\frac{(\theta- \tilde{\mu}^{(j)}_{k}(t))^2}{(\tilde{\sigma}^{(j)}_{k}(t))^2} -\frac{1}{2}\sum_{j=1}^{N}W_{ij}\log 2\pi (\tilde{\sigma}^{(j)}_k(t))^2
\\
& = 
-\frac{1}{2}\left(\theta^2\sum_{j=1}^N \frac{W_{ij}}{(\tilde{\sigma}^{(j)}_{k}(t))^2} - 2\theta \sum_{j=1}^N \frac{W_{ij}\tilde{\mu}^{(j)}_k(t)}{(\tilde{\sigma}^{(j)}_{k}(t))^2}
\right.
\\
&\quad  \left.+ \sum_{j=1}^N\frac{(\tilde{\mu}^{(j)}_k(t))^2}{(\tilde{\sigma}^{(j)}_{k}(t))^2} \right) 
-\frac{1}{2}\sum_{j=1}^{N}W_{ij}\log 2\pi (\tilde{\sigma}^{(j)}_k(t))^2.
\end{align} 
By completing the squares we have
\begin{align}
&\frac{1}{(\sigma^{(i)}_k(t+1))^{2}} 
= \sum_{j=1}^N W_{ij} 
\frac{1}{(\tilde{\sigma}^{(j)}_k(t+1))^{2}}
\\
&= \sum_{j=1}^N W_{ij}\frac{1}{\sigma^{(j)}_k(t)}
+ \sum_{j=1}^N W_{ij}\frac{\eta \indicate{A^{(j)}_t = k}}{\sigma^2},
\end{align}
and
\begin{align}
&\frac{\mu^{(i)}_k(t+1)}{(\sigma^{(i)}_k(t+1))^{2}} 
= \sum_{j=1}^N W_{ij} \frac{\tilde{\mu}^{(j)}_k(t+1)}{(\tilde{\sigma}^{(i)}_k(t+1))^{2}}
\\
&= \sum_{j=1}^N W_{ij}\frac{\mu^{(i)}_k(t)}{(\sigma^{(i)}_k(t))^{2}} + \sum_{j=1}^N W_{ij}\frac{\eta Y^{(j)}_t\indicate{A^{(j)}_t = k}}{\sigma^2}.
\end{align}

\subsection{Facts}
\begin{fact}
\label{fact:beta_binom}
\begin{align}
F^{\text{Beta}}_{\alpha, \beta}(x) 
= 1 - F^{\text{Binom}}_{\alpha+\beta+1, x}(\alpha-1),
\end{align}
for all positive integers $\alpha, \beta$.
\end{fact}

\begin{fact}
\label{fact:beta_cdf1}
If $\alpha_2 < \alpha_1$, then $F^{\text{Beta}}_{\alpha_1, \beta}(y) < F^{\text{Beta}}_{\alpha_2, \beta}(y) $. If $\beta_2 < \beta_1$, then $F^{\text{Beta}}_{\alpha, \beta_2}(y) < F^{\text{Beta}}_{\alpha, \beta_1}(y)$.
\end{fact}

\begin{fact}
\label{fact:beta_cdf2}
Let $\mu  \in [0,1]$ and for $n_1$, $n_2$ large enough and $n_2 < n_1$. Then,
\begin{align}
F^{\text{Beta}}_{\mu n_2 + 1, (1-\mu)n_2 + 1}(y)
< 
F^{\text{Beta}}_{\mu n_1 + 1, (1-\mu)n_1 + 1}(y),
\end{align}
for all $y > \mu$.
\end{fact}

\begin{fact}[Lemma~2 in~\cite{7349151}]
\label{fact:W_sum_conv}
Given a strongly connected network, the stochastic matrix $W$ satisfies 
\begin{align}
\sum_{\tau = 1}^t \sum_{j=1}^N 
\left| W^{t-\tau}_{ij}-\frac{1}{N}\right|
\leq \frac{4 \log N}{1- \sigma_2(W)}.
\end{align}
\end{fact}


\begin{fact}[Chernoff-Hoeffding Bound]
\label{fact:chernoff}
Let $X_1, \ldots, X_n$ be independent 0-1 random variables with $\expe[X_i] = p_i$. Let $X = \frac{1}{n}\sum_{i=1}^n X_i$  and $\expe[X] = \frac{1}{n}\sum_{i=1}^n p_i = \mu$. Then, for any $0 <\epsilon < 1-\mu$, 
\begin{align}
\P(X > \mu + \epsilon) \leq e^{-nd(\mu+\epsilon , \mu)},
\end{align}
and for $0 < \epsilon < \mu$,
\begin{align}
\P(X < \mu - \epsilon) \leq e^{-nd(\mu - \epsilon , \mu)},
\end{align} 
where $d(a , b) = a\log\frac{a}{b}+(1-a)\log \frac{1-a}{1-b}$.
\end{fact}

\end{document}